\documentclass[runningheads]{llncs}
\usepackage{graphicx}
\usepackage{tikz}
\usepackage{comment}
\usepackage{amsmath,amssymb} % define this before the line numbering.
\usepackage{color}
\usepackage{epsfig}
\usepackage{booktabs}
\usepackage{multirow}
\usepackage{array}
\usepackage{subfigure}
\usepackage{setspace}
\usepackage{wrapfig}
\usepackage{arydshln}
\usepackage[symbol]{footmisc}
\usepackage{algorithm, algorithmic} 
\usepackage[colorlinks,linkcolor=red]{hyperref}

\begin{document}
% \linenumbers
\pagestyle{headings}
\mainmatter
\def\ECCVSubNumber{591}  % Insert your submission number here
\makeatletter
\newcommand{\specificthanks}[1]{\@fnsymbol{#1}}% Inserts a specific \thanks symbol
\makeatother

\title{Learning Open Set Network with Discriminative Reciprocal Points} % Replace with your title

\begin{comment}
\titlerunning{ECCV-20 submission ID \ECCVSubNumber} 
\authorrunning{ECCV-20 submission ID \ECCVSubNumber} 
\author{Anonymous ECCV submission}
\institute{Paper ID \ECCVSubNumber}
\end{comment}
%******************

% \begin{comment}
\titlerunning{Learning Open Set Network with Discriminative Reciprocal Points}
\author{Guangyao Chen\inst{1} \and
Limeng Qiao\inst{1} \and
Yemin Shi\inst{1} \and % \orcidID{0000-0001-9024-7266}
Peixi Peng\inst{1} \inst{\ast}  \and 
Jia Li\inst{2,3} \and % \orcidID{0000-0002-4346-8696}
Tiejun Huang\inst{1,3} \and % \orcidID{0000-0002-4234-6099}
Shiliang Pu\inst{4} \and 
Yonghong Tian\inst{1,3} \thanks{Corresponding author}  % \orcidID{0000-0002-2978-5935} 
}

\authorrunning{G. Chen et al.}
% School of Electronics Engineering and Computer Science, Peking University \and 
\institute{
Department of Computer Science and Technology, Peking University \and
State Key Laboratory of Virtual Reality Technology and Systems, SCSE, Beihang University \and 
Peng Cheng Laborotory, Shenzhen, China \and 
Hikvision Research Institute, Hangzhou, China \\
\email{\{gy.chen, qiaolm, pxpeng, yhtian\}@pku.edu.cn}
}
% \end{comment}
%******************
\maketitle

\begin{abstract}
	Open set recognition is an emerging research area that aims to simultaneously classify samples from predefined classes and identify the rest as 'unknown'. In this process, one of the key challenges is to reduce the risk of generalizing the inherent characteristics of numerous unknown samples learned from a small amount of known data. In this paper, we propose a new concept, \emph{Reciprocal Point}, which is the potential representation of the extra-class space corresponding to each known category. The sample can be classified to known or unknown by the otherness with reciprocal points. To tackle the open set problem, we offer a novel open space risk regularization term. Based on the bounded space constructed by reciprocal points, the risk of unknown is reduced through multi-category interaction. The novel learning framework called \textbf{R}eciprocal \textbf{P}oint \textbf{L}earning (RPL), which can indirectly introduce the unknown information into the learner with only known classes, so as to learn more compact and discriminative representations. Moreover, we further construct a new large-scale challenging aircraft dataset for open set recognition: \textbf{Air}craft 300 (Air-300). Extensive experiments on multiple benchmark datasets indicate that our framework is significantly superior to other existing approaches and achieves state-of-the-art performance on standard open set benchmarks.
\end{abstract}

\section{Introduction}

\begin{figure}[!tb]
	\centering
	\includegraphics[width=\linewidth]{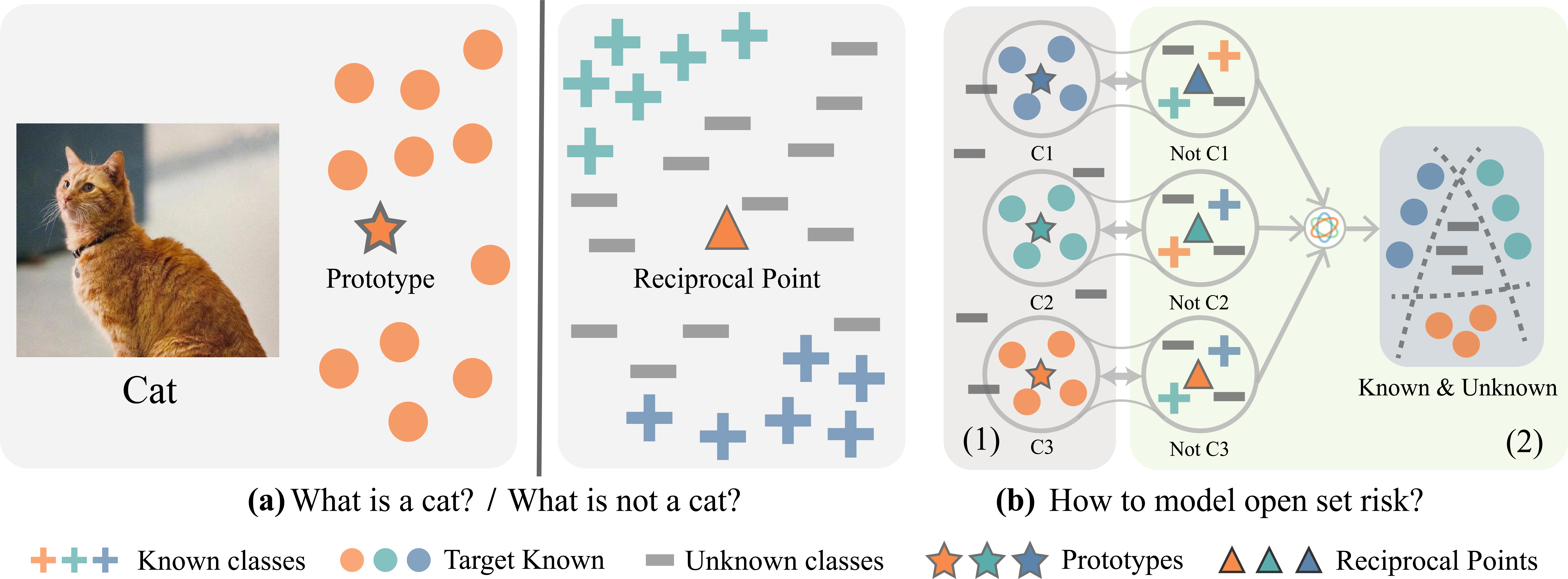}
	\caption{(a): How to identify a cat in open set problem? Most methods focus on learning the representative potential feature of cats as prototypes. In contrast, \emph{Reciprocal Points}, representative potential feature of \emph{non-cat}, identify the cat by otherness. (b): (1) Focusing only on known information, it is hard to reduce open space risk without unknown information. (2) The \emph{extra-class} space represented by \emph{Reciprocal Points} can bring more connections with unknown.}
	\label{fig:motivation}
\end{figure}

Recent error rate for classification task by Deep Convolutional Neural Network has surpassed the human-level performance \cite{he2015delving}. Such a significant improvement promotes its attempts in real-world applications. However, a robust recognition system should not only be able to identify all test instances of the seen or known classes, but also be able to handle unknown samples or novel events that have not been seen. Recent works on open set recognition \cite{scheirer2013toward} have formalized the process for performing recognition in the setting that labels the input as one of the known classes or as unknown. Apart from returning the most likely category, the robust recognition system must also support the rejection of unseen/unknown. An open set scenario has classes in testing phase that are not seen during training, which brings great difficulties to solve open set recognition. 

In the majority of deep neural networks, the last layer for multiclass recognition is the SoftMax function which produces a probability distribution over the known classes. However, the SoftMax layer brings a significant difficulty to open set recognition because of its closed nature. Bendale \emph{et al} \cite{bendale2016towards} improves SoftMax to OpenMax to estimate the probability of unknown classes but does not optimize for unknown classes during the training phase. Potential solutions to the open set recognition should optimize for unknown classes, as well as the known classes \cite{scheirer2013toward}. More knowledge of unknown information can facilitate reducing of open space risk. Many methods \cite{ge2017generative,neal2018open,oza2019c2ae,yoshihashi2018classification} utilize generative models for generating samples as unknown. However, Nalisnick \emph{et al} \cite{nalisnick2018deep} finds the instability of the mainstream generation models to the open set problem. How to model open space risk with only known training data is still an urgent problem to be solved. For example, as shown in Fig.\ref{fig:motivation}(a), how to identify a cat in open set problem. Most classification methods pay attention to \emph{"what is a cat?"}. They look for a more representative feature of the cat. Yang \emph{et al} \cite{yang2018robust} proposes convolutional prototype learning (CPL) to solve the multiclass recognition problem in the open world. For neural networks, each known class $k$ exists in its own way. Although focusing only on known samples can reduce the intra-class distance, it is still inevitable that it will bring open space risks because the dark information in the unknown space cannot be considered at all. In addition, for each known category $k$, unknown samples obviously belong to the space of non-$k$ and their features should be more similar to the representation of non-$k$ correspondingly for neural networks, which means that the corresponding unknown information is more implicit in each non-$k$ embedding space. Therefore, we further focus on \emph{"what is not a cat?"} to learn representations in the form of one more latent vector of non-cat data. The new representation is called \emph{Reciprocal Point}, which can bring more connections with unknown by exploring each extra-class space.

\begin{figure}[!tb]
	\centering
	\includegraphics[width=\linewidth]{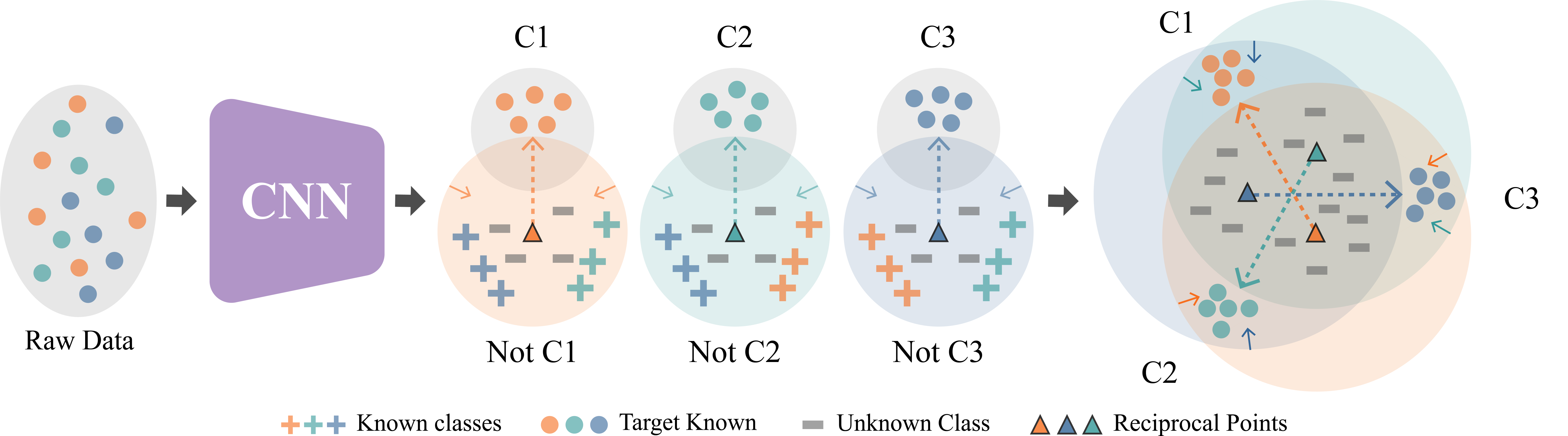}
	\caption{An overview of the proposed Reciprocal Point Learning (RPL) approach to open set recognition. The extra-class embedding space of each known class is limited to a bounded range through the reciprocal point and the learnable margin. With interaction of multiple classes, the known classes are pushed to the periphery of the feature space, and the unknown classes are limited in the bounded space.}
	\label{fig:overview}
\end{figure}

Based on the exploration of the extra-class space, we propose a novel framework, termed as \textbf{R}eciprocal \textbf{P}oint \textbf{L}earning (RPL), which can effectively reduce the open space risk by bounded space while learning reciprocal points of each known class, and the whole framework of the proposed method is shown in Fig. \ref{fig:overview}. Specifically, we extract the feature of each sample with a deep Convolutional Neural Network model. Since the reciprocal point is the learnable representation of the extra-class space, we classify the input by the otherness between the embedding feature and the reciprocal point. Then we extend the model to capture the risk of the unknown based on reciprocal points by a novel open space risk regularization term. The extra-class embedding space of each known class is limited to a bounded range through the reciprocal point and the learnable margin. When multiple classes interact with each other in the training stage, all known classes are not only pushed to the periphery of the space by the corresponding reciprocal point but also pulled in a certain bounded range by other reciprocal points. Finally, the embedding space of the network is limited to a bounded range. All known classes are distributed around the periphery of the embedding space, and the unknown samples are limited to the internal bounded space. The bounded domain by RPL can prevent the neural network from generating arbitrarily high confidence for unknown \cite{hein2019relu}. Although only known samples are available during the training stage, the interval between known and unknown classes is separated by reciprocal points indirectly. The advantage of RPL is that it can shrink unknown space while considering the known space classification and form a good bouned feature space distribution in which the magnitudes of unknown samples are lower than those of known samples.

Moreover, in order to facilitate the research of open set recognition in the real visual world, we further construct a new large-scale aircraft dataset from the Internet: \textbf{Air}craft 300 (Air-300), which contains 320,000 annotated colored images from 300 different classes in total. Note that each category contains 100 images at least, and a maximum of 10,000 images, which leads to the long tail distribution of our proposed dataset. Compared with the existing benchmark datasets, the tailored Air-300 dataset maintains a long tail distribution to simulate the real visual world, and can also be utilized to perform fine-grained classification and object recognition.

To summarize our contributions: (1) we propose a new concept, \emph{Reciprocal Point}, which is the potential representation of the extra-class space corresponding to each known category. (2) Meanwhile, we offer a novel framework for learning open set network. Through introducing unknown information by reciprocal points, the neural network can learn more compact and robust feature space, and separate known space and unknown space effectively. (3) Then, we further introduce a new challenging large-scale dataset with long-tailed distribution, Air-300, which is a new dataset to tailor for open set recognition. (4) The results on multiple benchmark datasets show that our approach significantly outperforms other existing state-of-the-art deep open set classifiers.

\section{Related Work}

\subsubsection{Open Set Recognition}
Inspired by a classifier with rejection option \cite{bartlett2008classification,da2014learning,yuan2010classification}, Scheirer \emph{et al.} \cite{scheirer2013toward} define the open set recognition problem for the first time and propose a base framework to perform training and evaluation. The deep neural network has achieved great success in many areas and is introduced to open set recognition by Bendale \emph{et al.}  \cite{bendale2016towards}. They prove threshold on SoftMax probability does not yield a robust model for open set recognition. Openmax \cite{bendale2016towards} detectes unknown classes by modeling distance of activation vectors, which are from the mean penultimate vector of each class with EVT. Ge \emph{et al} \cite{ge2017generative} propose G-Openmax, a direct extension of Openmax, which trains deep neural networks with unknown samples generated by generative models. However, this method does not extend to tasks that generate modeling difficulties. Yoshihashi \emph{et al.} \cite{yoshihashi2018classification} addresses deep reconstruction-based representation learning in open set recognition by adding image reconstructed as a regularizer. However, this method introduces auxiliary decoder network for training, which brings additional overhead to training.

\subsubsection{Out-of-Distribution Detection}
Based on the concern for safety of AI systems, the detection of out-of-distribution (OOD) examples, one of the concerns regarding the distribution shift, is first introduced by Hendrycks \emph{et al.} in \cite{hendrycks2016baseline}. OOD is the detection of samples that do not belong to the training set but can appear during testing \cite{hendrycks2016baseline}. Several works \cite{hendrycks2016baseline,liang2017enhancing,lee2017training,hendrycks2018deep} seek to address these problems by giving deep neural network classifiers a means of assigning anomaly scores to inputs, which are used for detecting OOD. Hendrycks \emph{et al.} \cite{hendrycks2016baseline} demonstrated that anomalous samples had a lower maximum softmax probability than in-distribution samples based on a deep, pre-trained classifier. Liang \emph{et al.} \cite{liang2017enhancing} propose ODIN to allow more effective detection by using temperature scaling and adding small perturbations to the input.  Similar with G-Openmax, De Vries \emph{et al.} \cite{hendrycks2018deep} utilize generative models for generating most effective samples from out-of-distribution and deriving a new OOD score form this branch. Hendrycks \emph{et al.} \cite{hendrycks2018deep} propose Outiler Exposure, which uses an auxiliary dataset in order to teach the network better representations for anomaly detection. Detecting out-of-distribution (OOD) is related to the rejection of unknown classes in open set recognition (OSR). They all are studying separating in-distribution (known) and out-of distribution (unkown) \cite{hendrycks2016baseline,scheirer2013toward}. 

\subsubsection{Prototype Learning}
Prototypes are learnable representations in the form of one or more latent vector per class. Wen \emph{et al.} \cite{wen2016discriminative} proposed a center loss to learn centers for deep features of each identity and used the centers to reduce intra-class variance. It can enhance the discrimination power of features. Yang \emph{et al.} in \cite{yang2018robust} propose Convolutional Prototype Learning (CPL) for robust classification. Moreover, a Generalized CPL (GCPL) with prototype loss was proposed as a regularization to improve the intra-class compactness of the feature representation. However, center loss and GCPL only explicitly encourage intra-class compactness. Conversely, the reciprocal point limits the embedding space outside each known class to bounded space. Known and unknown classes are separated by maximizing otherness between known and corresponding reciprocal points.

\section{Reciprocal Point Learning}

\subsection{Preliminaries}
\label{notion}
\noindent
We first establish preliminaries related to open set learning, following which we formally formulate the proposed idea. Suppose we are given a set of $n$ labeled samples $\mathcal{D}_\mathcal{L} = \lbrace (x_1, y_1), \dots, (x_n, y_n) \rbrace$ where $y_i \in \lbrace 1, \dots, N \rbrace$ is the label of $x_i$ and a larger amount of test data $\mathcal{D}_\mathcal{T}=\lbrace t_1, \dots, t_{u} \rbrace$ where the label of $t_i$ belong to $\lbrace 1, \dots, N \rbrace \cup \lbrace N+1, \dots, N+U \rbrace $ and $U$ is the number of unknown classes in actual scenarios. We then denote the embedding space of category $k$ as $\mathcal{S}_k$ and its corresponding \textbf{open space} as $\mathcal{O}_k = \mathbb{R}^d-\mathcal{S}_k$, where $\mathbb{R}^d$ is the $d$-dimensional full space. In order to formalize and then manage open space risk finely, we further address the positive open space from other known classes as $\mathcal{O}^{pos}_k$ and the remaining infinite unknown space as negative open space $\mathcal{O}^{neg}_k$, namely $\mathcal{O}_k = \mathcal{O}^{pos}_k \cup \mathcal{O}^{neg}_k$. 

\textbf{The Open Set Recognition Problem.} For simplicity, we will first introduce the open set recognition of a single class and then extend the entire learning process to a multiclass form. Given the labeled data $\mathcal{D}_\mathcal{L}$ with $N$ known classes, let samples from category $k$ be positive training data $\mathcal{D}^k_\mathcal{L}$ (space $\mathcal{S}_k$), samples from other known classes be negative training data $\mathcal{D}^{\neq k}_\mathcal{L}$ (space $\mathcal{O}^{pos}_k$), and samples from $\mathbb{R}^d$ but except $\mathcal{D}_\mathcal{L}$ be potential unknown data $\mathcal{D}_\mathcal{U}$ (space $\mathcal{O}^{neg}_k$). Let $\psi^k: \mathbb{R}^d \mapsto  \lbrace 0, 1 \rbrace $ be a binary measurable prediction function, mapping embedding $x$ to the label $y$. For 1-class open set recognition problem, the overall goal is to optimize a discriminant binary function $\psi_k$ by minimizing the expected error $\mathcal{R}^k$ as:
\begin{equation}
\mathop{\arg\min}_{\psi_k} \lbrace \mathcal{R}^k \mid         
   \mathcal{R}_\epsilon(\psi_k, \mathcal{S}_k \cup \mathcal{O}^{pos}_k) + \alpha \cdot
   \mathcal{R}_o(\psi_k, \mathcal{O}^{neg}_k) \rbrace
\label{Eqn:1}
\end{equation}
where $\alpha$ is a positive regularization parameter, $\mathcal{R}_\epsilon$ is the empirical classification risk on known data and $\mathcal{R}_o$ is the \textbf{open space risk} function to measure the uncertainty of labeling the unknown samples as the class of interest or as unknown, with further formulating as a non-zero integral function on space $\mathcal{O}^{neg}_k$. 

Afterwards we identify open set recognition problem by integrating multiple binary classification tasks (\textit{one} \textit{vs.} \textit{rest}) into a multiclass recognition problem (see Fig.\ \ref{fig:overview}), by summarizing the expected risk defined in the Eq.\eqref{Eqn:1} category by category, i.e.\ $\mathop{\sum}\nolimits_{k=1}^N \mathcal{R}^k$, which leads to the following formulation as:
\begin{equation}
\mathop{\arg\min}_{f \in \mathcal{H}} \ \lbrace 
\mathcal{R}_\epsilon(f, \mathcal{D}_L) + \alpha \cdot 
\mathop{\sum}\nolimits_{k=1}^N\mathcal{R}_o(f, \mathcal{D}_U) \rbrace
\label{Eqn:2}
\end{equation}
where $f: \mathbb{R}^d \mapsto \mathbb{N}$ is a measurable multiclass recognition function (see more details of derivation in supplementary material). According to the Eq.\ \eqref{Eqn:2}, solving the open set recognition problem is equivalent to minimize the combination of the empirical classification risk on labeled known data and the open space risk on potential unknown data simultaneously, over the space of allowable recognition functions, which  leads to be more distinguishable between known and unknown spaces. Further, we propose reciprocal points for both closed set classification and reducing the open space risk on $\mathcal{D}_U$ during training.

\subsection{Reciprocal Points for Classification}

The \textbf{reciprocal points} for category $k$ are denoted as $ \mathcal{P}^k = \{ p^k_i | i=1, \dots, M \} $, where $M$ is the number of reciprocal points for each class and $ \mathcal{P}^k $ can be regarded as the set of latent representations for the sub-dataset $\mathcal{D}^{\neq k}_\mathcal{L} \cup \mathcal{D}_\mathcal{U}$ (space $ \mathcal{O}_k $). Unlike the learning driven by prototype or center loss, we agree the fact that samples from the classes in $ \mathcal{O}_k $ are more similar to the classificatory reciprocal points $ \mathcal{P}^k $ than samples from  $ \mathcal{S}_k $, which can be formulated as:
\begin{equation}
    \forall d \in \zeta(\mathcal{P}^k, \ \mathcal{D}^k_L), \ 
	\max(\zeta(\mathcal{P}^k, \ \mathcal{D}^{\neq k}_L \cup \mathcal{D}_U))
	\ \leq \ d
	\label{Eqn:r-point}
\end{equation} 
where $\zeta(\cdot, \ \cdot)$ calculates a set of distances of all samples between two sets and max($\cdot$) is a maximal function.

In fact, the most intuitive goal of open set learning is usually to separate known and unknown spaces as much as possible. To achieve this, for any category k, we propose to separate the two mutually exclusive space by maximizing the distance between the reciprocal points of the category and its corresponding known samples. Specifically, we optimize a set of $d$-dimensional representations $\mathcal{P}^k$, or reciprocal points, of each class through an embedding function $f_{\theta}$ with learnable parameters $\theta$. Given the sample $x$ and classificatory reciprocal points $\mathcal{P}_k$, we calculate the distance between them with the formulation as:
\begin{equation}
	d(f_{\theta}(x), \mathcal{P}^k) = \frac{1}{M}\sum\nolimits^{M}_{i=1} ||f_{\theta}(x) - p^k_i||^2_2
\end{equation} 
Afterwards, based on the distance we propose above, our framework estimates the otherness between the embedding feature $f_{\theta}(x)$ and the reciprocal points $\mathcal{P}^k$ of all known classes to determine which category it belongs to. Actually, following the nature of reciprocal points, the probability of sample $x$ belongs to category $k$ is proportional to the otherness between $f_{\theta}(x)$ and the reciprocal points of category $k$, i.e. $p(y=k|x) \propto d(f_{\theta}(x), \mathcal{P}^k)$, which means the greater the distance between $f_{\theta}(x)$ and $ \mathcal{P}^k $, the greater the probability that $x$ will be assigned with label $k$. According to the sum-to-one property, we embody the final probability as:
\begin{equation}
	p(y=k|x, f_{\theta}, \mathcal{P}) = \frac{e^{\gamma d(f_{\theta}(x), \  \mathcal{P}^k)}}{\sum\nolimits_{i=1}^{N}{e^{\gamma d(f_{\theta}(x), \ \mathcal{P}^i)}}}
\end{equation}
where $\gamma$ is a hyper-parameter that controls the hardness of probability assignment. Learning proceeds by minimizing the reciprocal points classification loss based negative log-probability of the true class k via SGD as:
\begin{equation}
	\mathcal{L}_c(x; \theta, \mathcal{P}) = - \ log \ p(y=k|x, f_{\theta}, \mathcal{P})
	\label{Eqn:loss}
\end{equation}
which can be further seen as integrating multiple binary classification tasks to solve the multiclass open set recognition problem. Through minimizing Eq.\ \eqref{Eqn:loss}, on the one hand, maximizing the otherness between known data $\mathcal{D}_L$ and reciprocal points set $\mathcal{P}$ has a facilitating effect on maximizing the interval between closed space $\mathcal{S}_k$ and open space $ \mathcal{O}_k $, which is consistent with our initial goal. On the other hand, corresponding to $ \mathcal{R}_\epsilon(f, \mathcal{D}_L) $ in the Eq.\ref{Eqn:2}, the reciprocal points classification loss reduces the empirical classification risk through the reciprocal points.

\subsection{Reducing Open Space Risk}

In solving the open set recognition problem, in addition to utilizing the reciprocal points loss to manage the empirical risk, we also further reduce the open space risk $\mathcal{R}_o(f, \mathcal{D}_U) $ in Eq. \ref{Eqn:2} with introducing a regularization term. According to the notion of open set in Sec. \ref{notion}, each particular category k has positive open space $\mathcal{O}_k^{pos}$ and infinite negative open space $\mathcal{O}_k^{neg}$ respectively. For multiclass open-set recognition scenarios, we further union multiple class-wise open spaces into a global open space $\mathcal{O}_G$ as:
\begin{equation}
	\mathcal{O}_G = \bigcap\nolimits^{N}_{k=1} (\mathcal{O}_k^{pos} \cup \mathcal{O}_k^{neg} )
\end{equation}
where we would restrict the total open space risk for all known classes. Moreover, based on the principle of maximum entropy, for an unknown sample $x_u$ without any prior information, a well-trained closed-set discriminant function tends to assign known labels to $x_u$ with equal probability, which leads to the deep neural networks usually embed unknown samples into the inside of known spaces rather than random positions in the full space. This phenomenon is also consistent with the observation of \cite{dhamija2018reducing} and our visualization results shown in Fig. \ref{fig:embedding}. 

Obviously, the idea of which wants to manage open space risk by restricting the open space to a bounded space directly is almost impossible because the open space contains a large number of potentially unknown samples. However, considering space $\mathcal{S}_k$ and $\mathcal{O}_k$ are complementary with each other, we indirectly bound the open space risk by constraining the distance between samples from $\mathcal{S}_k$ and reciprocal points $\mathcal{P}^k$ to a certain extent as:
\begin{equation}
	\mathcal{L}_o(x; \theta, \mathcal{P}^k, R^k) = \frac{1}{M} \cdot \sum^{M}_{i=1}{|| d(f_{\theta}(x), p^k_i) - R^k ||^2_2}
	\label{Eqn:regularization}
	\end{equation}
where $R^k$ is the learnable margin in our framework. Specifically, minimizing the Eq. \eqref{Eqn:regularization} is equivalent to making $R^k$ and $\zeta(\mathcal{D}^k_L, \ \mathcal{P}^k)$ in Eq. \eqref{Eqn:r-point} as close as possible, so that the following formula is established as: 
\begin{equation}
% \forall r^k_i \in R^k, \ 
\max(\zeta(\mathcal{D}^{\neq k}_L \cup \mathcal{D}_U, \ \mathcal{P}^k)) 
\ \leq \ \
R^k
% r^k_i
\label{Eqn:reg}
\end{equation} 

Considering bounded space $\mathcal{B}(\mathcal{P}^k, R^k)$ with the reciprocal points $\mathcal{P}^k$ as centers and $R^k$ as its corresponding intervals, in order to separate known and unknown space, we further utilize these bounded spaces to approximate the global unknown space $\mathcal{O}_G$ as much as possible. As a result, calculating the regularization loss with Eq. \eqref{Eqn:regularization} can be viewed as managing open space risk $\mathcal{R}_o(f, \mathcal{D}_U)$ in Eq. \eqref{Eqn:loss}. 

\begin{algorithm}[!tb]
	\fontsize{8}{8}
	% \footnotesize
	% \small
	\renewcommand{\algorithmicrequire}{\textbf{Input:}}
	\renewcommand{\algorithmicensure}{\textbf{Output:}}
	\caption{The reciprocal point learning algorithm.}
	\label{alg:1}
	\begin{algorithmic}[1]
		\REQUIRE Training data $\{ x_i \}$. Initialized parameters $\theta$ in convolutional layers. Parameters $\mathcal{P}$ and $R$ in loss layers, respectively. Hyperparameter $ \lambda, \gamma $ and learning rate $ \mu $. The number of iteration $t \leftarrow 0$.
		\ENSURE The parameters $\theta$, $\mathcal{P}$ and $R$.
		\STATE while \textit{not converge} do
		\STATE  \qquad $t \leftarrow t + 1$.
		\STATE  \qquad Compute the joint loss by $ \mathcal{L}^t = \mathcal{L}_c^t + \lambda \cdot \mathcal{L}_o^t $.
		\STATE  \qquad Compute the backpropagation error $\frac{\partial \mathcal{L}^t}{\partial x^t}$ for each $i$ by $\frac{\partial \mathcal{L}^t}{\partial x^t} = \frac{\partial \mathcal{L}_c^t}{\partial x^t} + \lambda \cdot \frac{\partial \mathcal{L}_o^t}{\partial x^t}$.
		\STATE  \qquad Update the parameters $\mathcal{P}$ by $\mathcal{P}^{t+1} = \mathcal{P}^{t} - \mu^t \cdot \frac{\partial \mathcal{L}^t}{\partial \mathcal{P}^t} = \mathcal{P}^{t} - \mu^t \cdot (\frac{\partial \mathcal{L}_c^t}{\partial \mathcal{P}^t} + \lambda \cdot \frac{\partial \mathcal{L}_o^t}{\partial \mathcal{P}^t})$.
		\STATE  \qquad Update the parameters $R$ by $R^{t+1} = R^{t} - \mu^t \cdot \frac{\partial \mathcal{L}^t}{\partial R^t} = R^{t} - \lambda \cdot \mu^t \cdot \frac{\partial \mathcal{L}_o^t}{\partial R^t}$.
		\STATE \qquad Update the parameters $\theta$ by $\theta^{t+1} = \theta^{t} - \mu^t \cdot \frac{\partial \mathcal{L}^t}{\partial \theta^t} = \theta^{t} - \mu^t \cdot (\frac{\partial \mathcal{L}_c^t}{\partial {\theta}^t} + \lambda \cdot \frac{\partial \mathcal{L}_o^t}{\partial \theta^t})$.
		\STATE end while
	\end{algorithmic}
\end{algorithm}

\subsection{Learning Open Set Network} 

Finally, the overall loss function of reciprocal points learning combines the empirical classification risk and reducing open space risk as:
\begin{equation}
	\mathcal{L}(x; \theta, \mathcal{P}, R) = \mathcal{L}_c(x;\theta, \mathcal{P}) + \lambda \mathcal{L}_o(x; \theta, \mathcal{P}, R)
	\label{Eqn: final-loss}
\end{equation}
where $\lambda$ is a hyper-parameter of controlling the weight of reducing open space risk module and $\theta, \mathcal{P}, R$ represent the learnable parameters of our framework. As shown in Fig. \ref{fig:overview}, learning open set recognition problem with Eq. \eqref{Eqn: final-loss} results in pushing the known spaces to the periphery of global open space and then separating the two spaces as much as possible. In Algorithm \ref{alg:1}, we summarize the learning details of the open set network with joint supervision.

\textbf{How to detect unknown classes?} Samples from a class on which the network was not trained would have probability scores distributed across all the known classes \cite{dhamija2018reducing}. Similar to thresholding softmax scores in \cite{hendrycks2016baseline}, we agree that unknown samples have a closer distance with all reciprocal points than samples of known training classes based on a deep classifier. Therefore, the probability that test instance $x$ belongs to known classes is proportional to the distance between the test instance and the farthest reciprocal point corresponding to category $k$:
\begin{equation}
p(known|x) \propto \max_{k \in \lbrace 1, \dots, N \rbrace}{d(f(x), \mathcal{P}^k)}
\end{equation}
The models for open set problem focus on two key questions, \emph{\textbf{Q1}) what is a good score for open-set identification?} (i.e., identifying a class as known or unknown), and given a score, \emph{\textbf{Q2}) what is a good operating threshold for the model?} \cite{oza2019c2ae}. Furthermore, since how rare or common samples from unknown space is not known in the actual scenario, the approaches to open set recognition which require an arbitrary threshold or sensitivity for comparison is unreasonable \cite{neal2018open}. Thus, we utilize the difference between known and unknown distribution to measure the learned models' ability to detect unknown, which provides a calibration-free measure of detection performance. 

\section{Experiments}

\subsection{Experiments for Open Set Identification}

\begin{table*}[!tb]
	\caption{The AUROC results of open set identification. The second column shows the backbone for each method, including Encoder (E) or Decoder (D). The architecture used to infer the unknown class are highlighted in bold. Values other than RPL are taken from \cite{oza2019c2ae}. RPL++ means using GCPL to assist RPL training. Best performing methods are highlighted in bold.}
	\centering
	\scriptsize
	\label{tab:osdi}
	\begin{tabular}{cccccccc}
		\toprule
		Method & Backbone & MNIST & SVHN & CIFAR10 & CIFAR+10 & CIFAR+50 & TinyImageNet \\
		\midrule
		Softmax & \textbf{E} & 97.8	& 88.6	& 67.7	& 81.6	& 80.5	& 57.7 \\
		Openmax \cite{bendale2016towards} & \textbf{E} & 98.1	& 89.4	& 69.5	& 81.7	& 79.6	& 57.6 \\
		RPL	& \textbf{E} & 98.9 & 93.4 & 82.7 & 84.2 & 83.2 & 68.8 \\
		RPL++ & \textbf{E} & \textbf{99.3} & \textbf{95.1} & \textbf{86.1} & \textbf{85.6} & \textbf{85.0} & \textbf{70.2} \\
		\midrule
		G-OpenMax \cite{ge2017generative} & \textbf{E} + E + D & 98.4	& 89.6	& 67.5	& 82.7	& 81.9	& 58.0 \\
		OSRCI \cite{neal2018open} & \textbf{E} + E + D & 98.8 & 91.0 & 69.9 & 83.8 & 82.7 & 58.6 \\
		C2AE \cite{oza2019c2ae} & \textbf{E + D} & 98.9 & 92.2 & 89.5 & 95.5 & 93.7 & 74.8 \\
		RPL-WRN	& \textbf{E}(WRN) & \textbf{99.6} & \textbf{96.8} & \textbf{90.1} & \textbf{97.6} & \textbf{96.8} & \textbf{80.9} \\
		\bottomrule
	\end{tabular}
\end{table*}

\noindent
\textbf{Evaluation Metrics.} The evaluation protocol defined in \cite{neal2018open} is considered and Area Under the Receiver Operating Characteristic (AUROC) curve is used as evaluation metric. AUROC is a threshold-independent metric \cite{davis2006relationship}. It can be interpreted as the probability that a positive example is assigned a higher detection score than a negative example \cite{fawcett2006introduction}. Following the protocol in \cite{neal2018open}, we report the AUROC averaged over five randomized trials. 

\noindent
\textbf{Network Architecture.} Except for RPL-WRN, the encoder and decoder architecture for this experiment is same to the architecture used in \cite{neal2018open}. For a fair comparison, we trained RPL with a Wide-ResNet \cite{zagoruyko2016wide} with depth 40, width 4 and dropout 0, WRN-40-4. The parameters of WRN-40-4 are about $9M$, which is less than the networks of C2AE \cite{oza2019c2ae} and OSRCI \cite{neal2018open}. $ \lambda $ for RPL is set to 0.1 in the training stage. More details in the supplementary material.

\noindent
\textbf{Datasets.} Similar to \cite{oza2019c2ae}, we provide a simple summary of these protocols for each dataset. 
\begin{itemize}
	\item \textbf{MNIST}, \textbf{SVHN}, \textbf{CIFAR10}. For MNIST \cite{lecun1998gradient}, SVHN \cite{netzer2011reading} and CIFAR10 \cite{krizhevsky2009learning}, by randomly sampling 6 known classes and 4 unknown classes. 
	\item \textbf{CIFAR+10}, \textbf{CIFAR+50}. For CIFAR+$N$ experiments, 4 classes are sampled from CIFAR10 for training. $N$ non-overlapping classes are used as unknown, which are sampled from the CIFAR100 dataset \cite{krizhevsky2009learning}. 
	\item \textbf{TinyImageNet}. For experiments with TinyImageNet \cite{russakovsky2015imagenet}, 20 known classes and 180 unknown classes are randomly sampled for evaluation.
\end{itemize}

\noindent
\textbf{Result Comparisons.}
As shown in Table \ref{tab:osdi}, RPL outperforms other methods based on encoder in \cite{neal2018open} in all datasets. Noted that RPL with only using known training samples and a single encoder outperforms methods based on encoder and decoder in MNIST and SVHN. With less information and simple training method, RPL achieves better performance than OSRCI \cite{neal2018open} and G-OpenMax \cite{ge2017generative} using generation-based model on all datasets. Furthermore, we need fewer network parameters and a more simple strategy for training to identify unknown classes. RPL-WRN has similar parameters to C2AE model and performs significantly better than other recent state-of-the-art methods in SVHN, CIFAR, and TinyImageNet. It shows that open set network trained by RPL learns better scores for identifying known and unknown classes. 

\subsection{Experiments for Open Long-Tailed Recognition}

\noindent
\textbf{Datasets.}
In real visual world, the frequency distribution of visual classes is long-tailed, with a few common classes and many more rare classes \cite{liu2019large}. To simulate natural data distribution, we adopt ImageNet-LT \cite{liu2019large} and a new image datasets, Air-300, to evaluate our algorithm for open set identification.

\begin{figure*}[!tb]
	\centering
	\subfigure[]{
		\begin{minipage}[t]{0.32\linewidth}
			\centering
			\includegraphics[width=\linewidth]{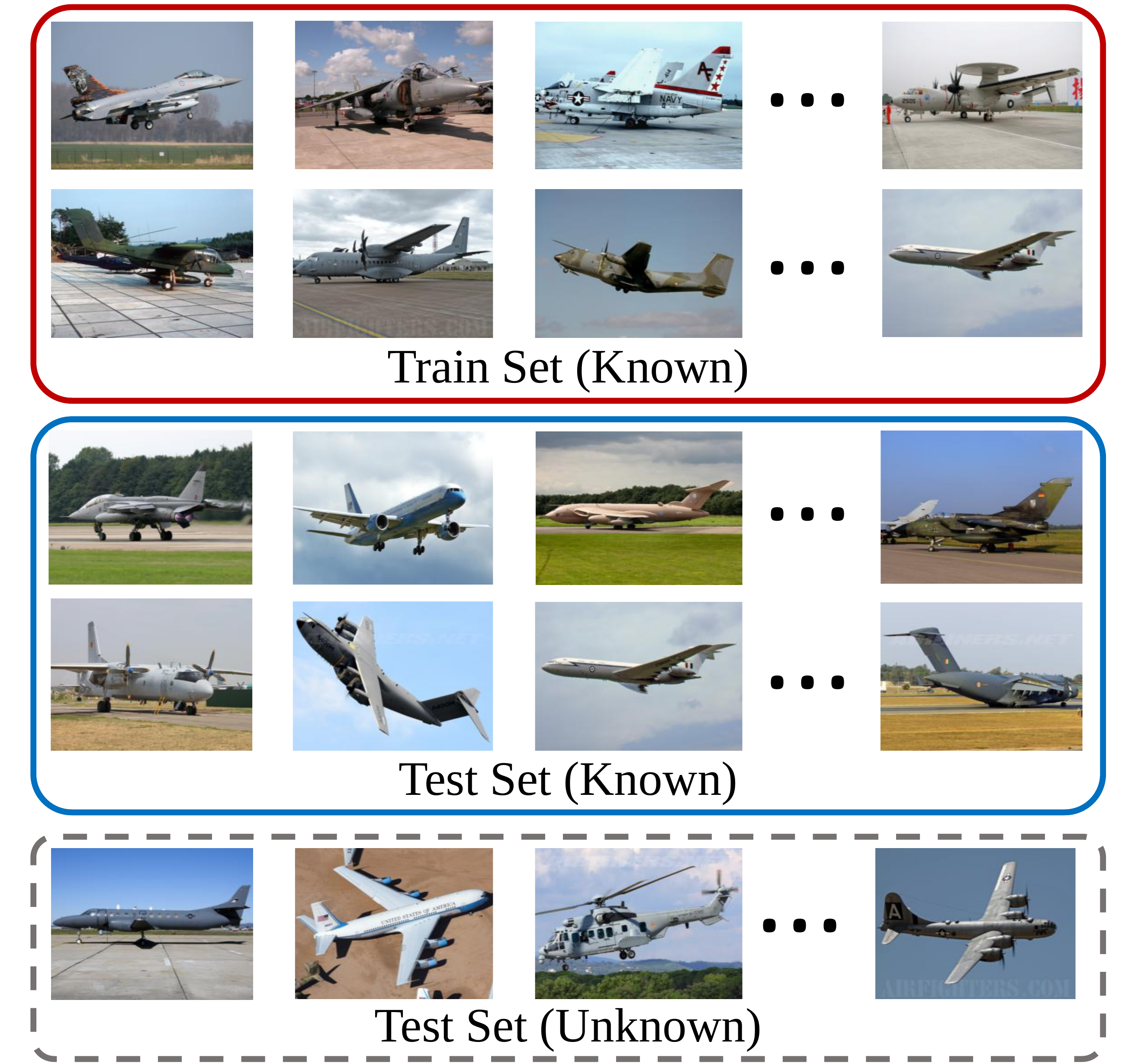}
			\label{fig:thumb}
		\end{minipage}
	}
	\subfigure[]{
		\begin{minipage}[t]{0.62\linewidth}
			\centering
			\includegraphics[width=\linewidth]{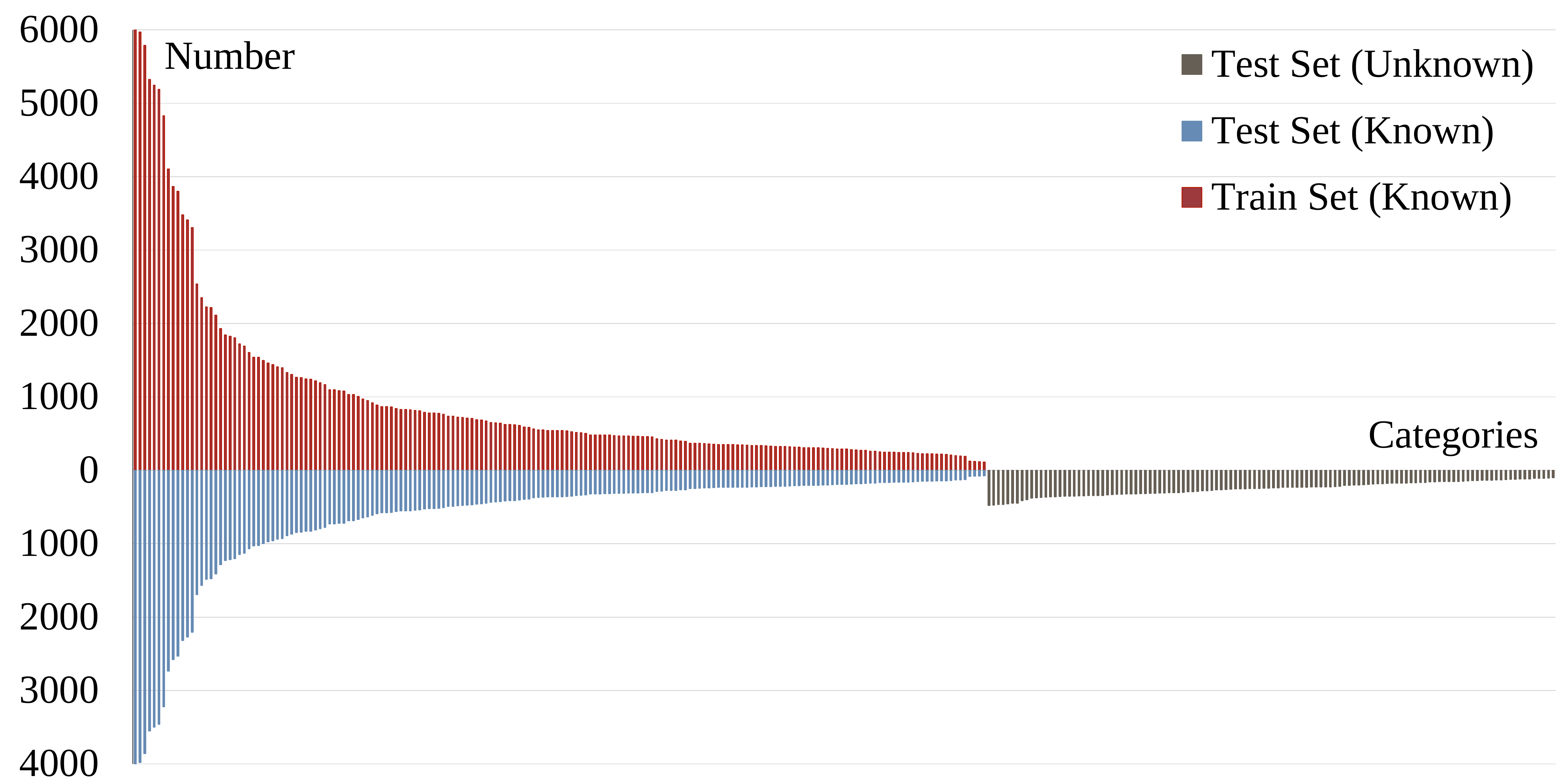}
			\label{fig:fine}
		\end{minipage}
	}
	\caption{(a) The overview of Air-300. (b) The number of images with classes.}
\end{figure*}
\textbf{ImageNet-LT.} ImageNet-LT is a long-tailed version of the original ImageNet-2012 \cite{deng2009imagenet}. It has 115.8K images from 1000 categories, with maximally 1280 images per class and minimally 5 images per class. The test set contains the open set based on the additional classes of images in ImageNet-2010.

\textbf{Aircraft 300 Dataset.} 
To facilitate research on open set recognition and evaluate the performance of different approaches, we further construct a new large-scale dataset from the Internet, \textbf{Air}craft \textbf{300} (Air-300), which contains 320,000 annotated colour images from 300 different classes in total, see thumbnails in Figure \ref{fig:thumb} briefly. Each category contains 100 images at least, and a maximum of 10,000 images, which leads to the long tail distribution as shown in Figure \ref{fig:fine}. According to the number of images in each category, we then divide all classes into two parts with 180 known classes for training and 120 novel unknown classes for testing respectively. Furthermore, we aggregate 300 different aircraft models into 10 super classes according to their roles, such as a bomber, fighter, helicopter and so on. As such, each image is annotated with a fine-grain label (the sub-class category) and a coarse-grain label (the superclasss category) by human. Note we also split these 10 superclasses into two parts for coarse-grain open set recognition. Compared with the existing benchmark datasets, the tailored Air-300 dataset is not only much larger and brings new challenges to open set recognition, but also can be utilized to perform fine-grain classification and object recognition respectively.

\noindent
\textbf{Evaluation Metrics.}
The AUROC and AUPR are adopted for evaluation, AUPR-Known and AUPR-Unknown denote the area under the precision-recall curve where known and unknown are specified as positives, respectively.

\noindent
\textbf{Network Architecture.}
All methods are trained based on WRN-40-4 for Air-300 and ResNet50 for ImageNet-LT \cite{liu2019large}. We optimize GCPL performance to best by adjusting its parameters. More details in the supplementary material.

\begin{table}[!tb]
	\caption{Test accuracy and open set identification of different methods on Air-300.}
	\centering
	\small
	\label{tab:olr}
	\begin{tabular}{llcccccccc}
		\toprule
		\multicolumn{2}{c}{\multirow{2}*{Method}} & \multicolumn{4}{c}{\textbf{Air-300} (WRN-40-4)} & \multicolumn{4}{c}{\textbf{ImageNet-LT} (ResNet50)} \\
		\cmidrule(r){3-6} \cmidrule(r){7-10}
		& & Acc & AUROC & AUPR-K & AUPR-U & Acc & AUROC & AUPR-K & AUPR-U \\
		\midrule
		\multicolumn{2}{c}{Softmax} & 85.7 & 77.1 & 85.8 & 48.0 & 37.8 & 53.3 & 45.0 & 60.5 \\
		\multicolumn{2}{c}{GCPL} & 84.5 & 79.1 & 88.0 & 61.3 & 37.1 & 54.5 & 45.7 & 61.9 \\
		\cdashline{1-10}
		\multirow{4}*{RPL} & $ \lambda = 0 $ & 84.6 & 64.6 & 78.5 & 43.6 & 38.7 & 55.0 & 46.0 & 62.5 \\
		& $ \lambda = 0.01 $ & 87.9 & 82.5 & 89.3 & 68.9 & \textbf{39.0} & 55.0 & 46.0 & 62.5 \\
		& $ \lambda = 0.1 $ & \textbf{88.8} & \textbf{83.1} & \textbf{89.3} & \textbf{70.2} & \textbf{39.0} & \textbf{55.1} & \textbf{46.1} & \textbf{62.7} \\
		& $ \lambda = 1 $ & 86.7 & 81.7 & 87.5 & 69.7 & 38.2 & 54.4 & 45.7 & 61.6 \\
		\cdashline{1-10}
		\multicolumn{2}{c}{RPL++} & \textbf{89.0} & \textbf{84.1} & \textbf{90.1} & \textbf{72.7} & \textbf{39.7} & \textbf{55.2} & \textbf{46.2} & \textbf{62.7} \\
		\bottomrule
	\end{tabular}
\end{table}

\noindent
\textbf{Result Comparisons.}
Table. \ref{tab:olr} shows the test accuracy of known classes and the performance of open set identification on Air-300 and ImageNet-LT. Under the $ \lambda = 0.1 $ settings, RPL has greatly improved in all evaluations compared to softmax and GCPL. For air-300, RPL not only improves the classification accuracy of closed set, but also is about 6\% higher than softmax in AUROC. ImageNet-LT has more categories and fewer training samples, which makes it very challenging for open set recognition. It is difficult to detect unknown when the accuracy of closed set is not so well, but RPL still improves the performance of classification and open set identification by nearly 2\%. It shows that RPL is effective for the open set problem of long tail distribution.

\subsection{Further Analysis}

% \subsubsection{Analysis of Reducing Open Space Risk.}
% \newline
\noindent
\textbf{Analysis of Reducing Open Space Risk.}

\begin{figure}[!tb]
	\subfigure[Softmax]{
		\begin{minipage}[t]{0.24\linewidth}
			\centering
			\includegraphics[width=\linewidth]{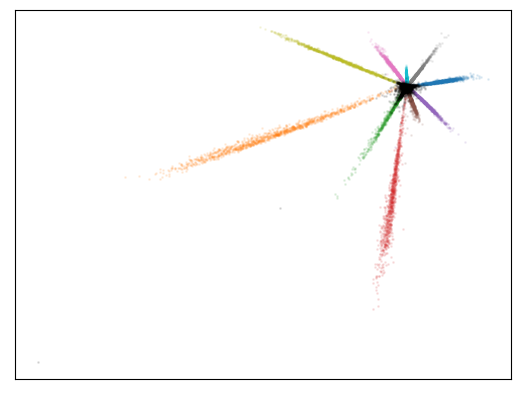}
			\includegraphics[width=\linewidth]{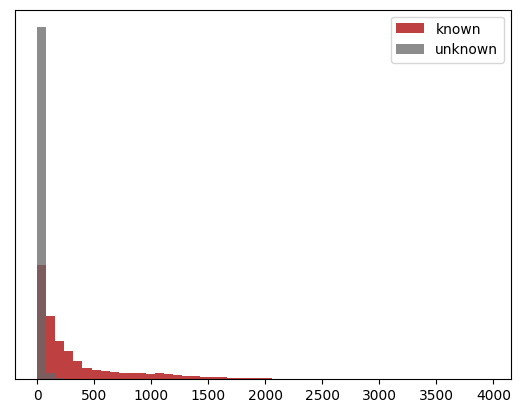}
			\label{fig:softmax}
		\end{minipage}%
	}
	\subfigure[RPL($\lambda = 0$)]{
		\begin{minipage}[t]{0.24\linewidth}
			\centering
			\includegraphics[width=\linewidth]{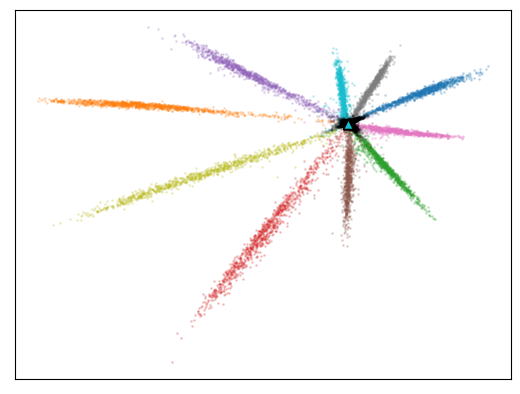}
			\includegraphics[width=\linewidth]{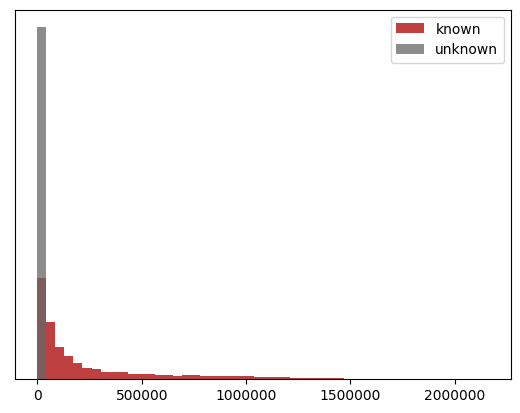}
			\label{fig:RPL_0}
		\end{minipage}%
	}%
	\subfigure[GCPL]{
		\begin{minipage}[t]{0.24\linewidth}
			\centering
			\includegraphics[width=\linewidth]{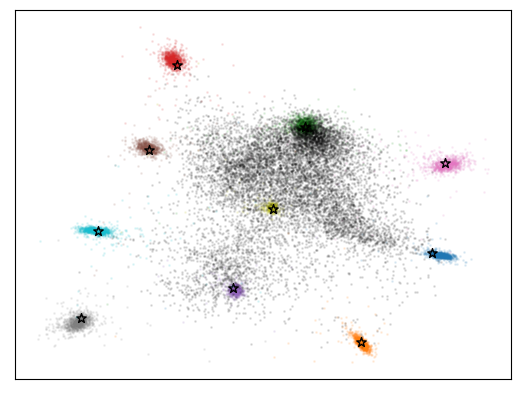}
			\includegraphics[width=\linewidth]{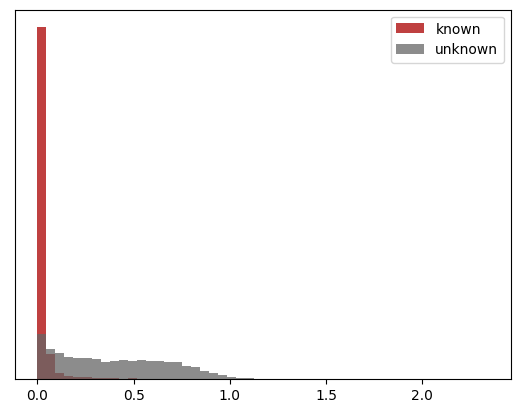}
			\label{fig:GCPL}
		\end{minipage}%
	}%
	\subfigure[RPL($\lambda = 0.1$)]{
		\begin{minipage}[t]{0.24\linewidth}
			\centering
			\includegraphics[width=\linewidth]{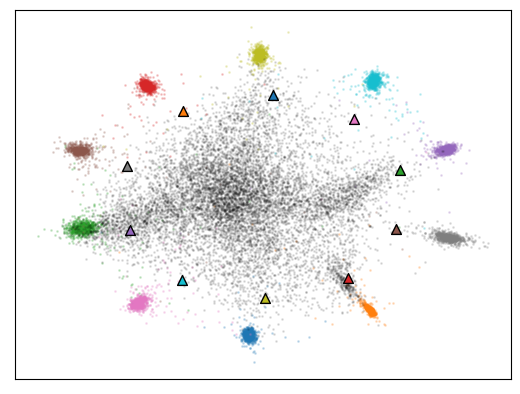}
			\includegraphics[width=\linewidth]{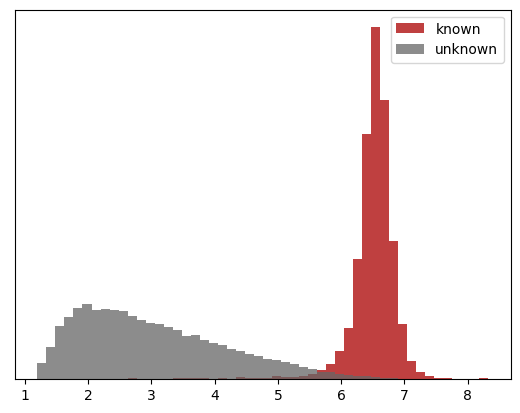}
			\label{fig:RPL}
		\end{minipage}%
	}
	\caption{(1) The first row is visualization in the learned feature space of MNIST as known and FashionMNIST as unknown. Gray shapes are data of unknown, and circles in color are data of known samples. Different colors represent different classes. Colored stars and triangles represent the prototypes and the reciprocal points learned of different known categories, respectively. (2) The second row is the maximum distance distribution between features and prototypes or reciprocal points.}
	\label{fig:embedding}
\end{figure}

\textbf{RPL vs. Softmax.}
Reciprocal points classification loss term (the first part of Eq. \ref{Eqn:2}) defines constraints for reciprocal point. However, similar to softmax, the learned representation is still linear separable only with this classification loss. See Fig.\ref{fig:softmax} and \ref{fig:RPL_0}, reciprocal points without $\mathcal{L}_o$ are learned to the origin, while RPL with $\mathcal{L}_o$ can achieve better distribution (shown in Fig. \ref{fig:RPL}). For softmax and RPL without $\mathcal{L}_o$, there is a significant overlap between the known and the unknown classes, so as to the neural network views unknown samples as some known classes with high confidence. In contrast, the whole feature space learned by RPL with $\mathcal{L}_o$ is in a limited range (1-6 in abscissa for unknowns and 5-8 in abscissa for knowns in Fig.\ref{fig:RPL}), which prevents high confidence in unknown classes.

\textbf{RPL vs. GCPL.} 
As shown in Fig.\ref{fig:GCPL}, GCPL used the prototypes to reduce intra-class variance. However, without considering the unknown, GCPL extends unknown classes to the whole feature space, resulting in a significant overlap with known. By introducing $\mathcal{L}_o$, different known categories are spread to the periphery of the space, while unknown categories are restricted to the interior. It can be seen that a clear gap is maintained between the two types of samples (known vs. unknown) in Fig.\ref{fig:RPL}. RPL improves the robustness of neural networks by preventing the misjudgment of the unknown class through the bounded restriction, thereby enhancing and stabilizing the classification of known categories.

\begin{figure}[!tb]
	\centering
	\subfigure[]{
		\begin{minipage}[t]{0.26\linewidth}
			\centering
			\includegraphics[width=\linewidth]{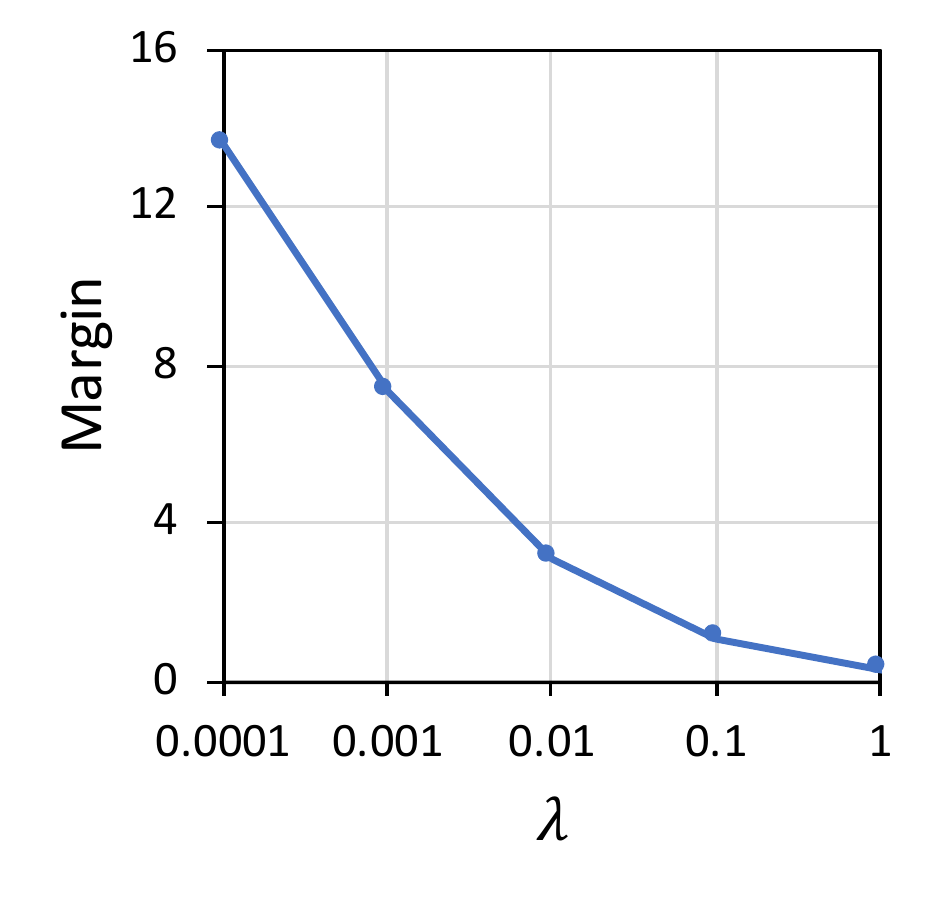}
			\label{fig:lamdbma}
		\end{minipage}%
		\begin{minipage}[t]{0.26\linewidth}
			\centering
			\includegraphics[width=\linewidth]{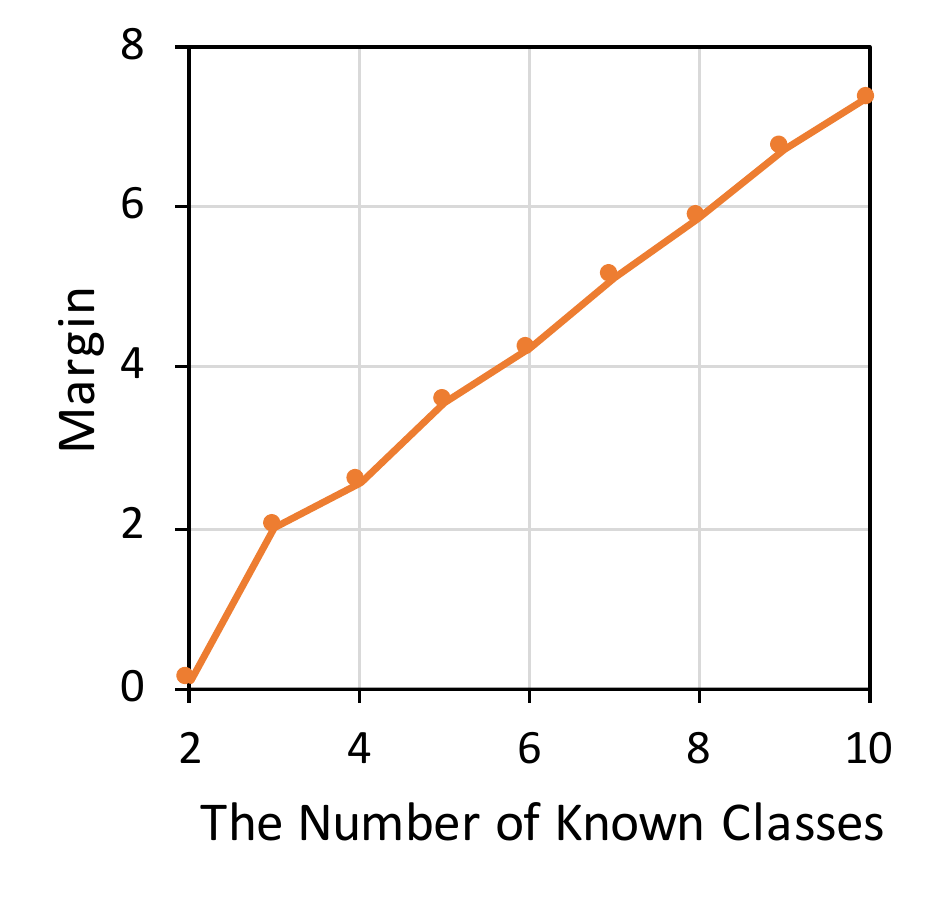}
			\label{fig:known}
		\end{minipage}%
	}%
	\subfigure[]{
		\begin{minipage}[t]{0.37\linewidth}
			\centering
			\includegraphics[width=\linewidth]{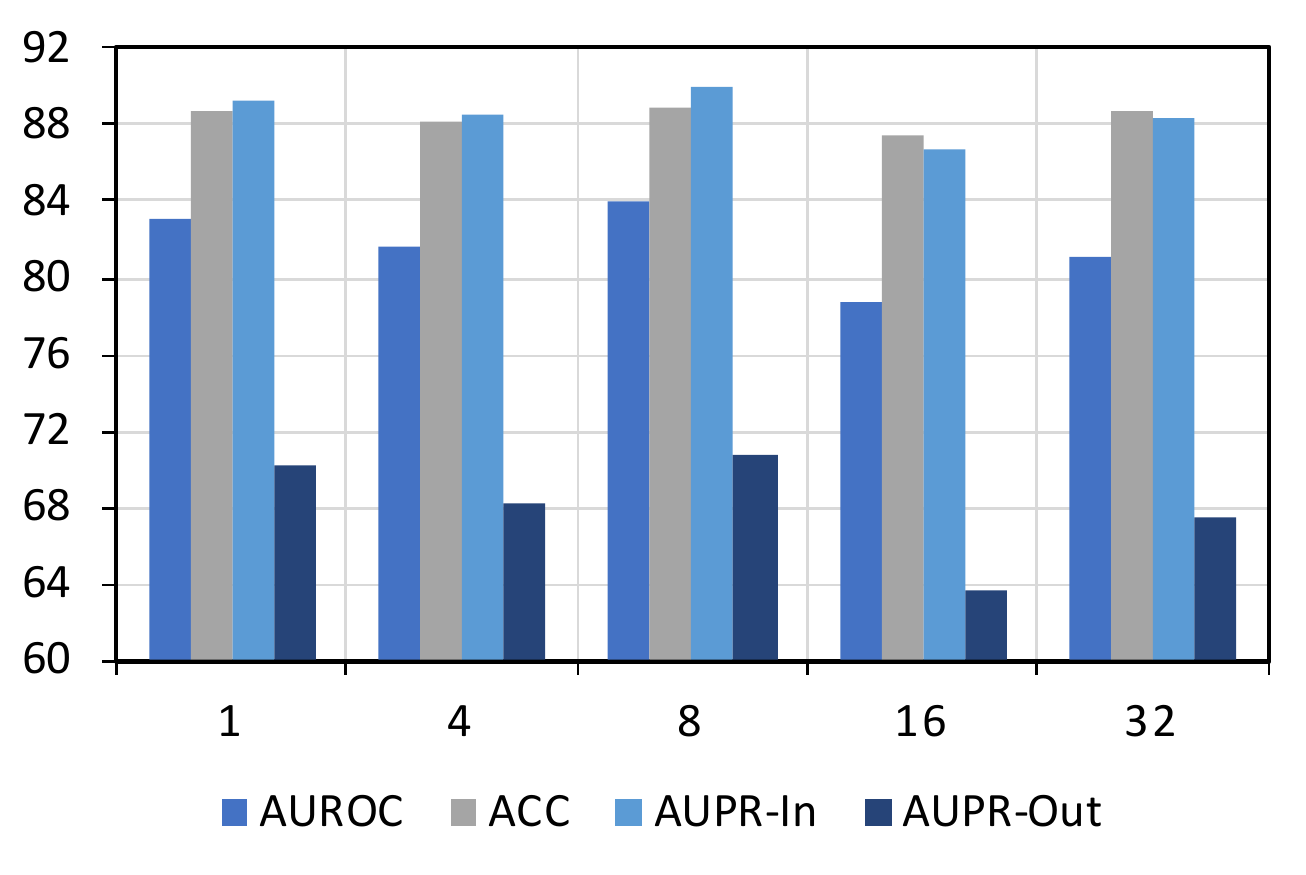}
			\label{fig:multi}
		\end{minipage}%
	}
	\caption{(a) The trend of margin with $\lambda$ and the number of known classes. (b) is the variation trend of testing performance with multiple reciprocal points for open set identification in Air-300. The abscissa represents the number of reciprocal points.}
	\label{fig:analysis}
\end{figure}

\textbf{Margin \& $\lambda$.} 
See Table. \ref{tab:olr}, the performance of RPL is related to the setting of $\lambda$. With the increase of $\lambda$, RPL has greatly improved on open set identification. However, when $\lambda$ is too big, the accuracy of known classes will decrease slightly, which is related to $\mathcal{L}_o$ limiting the size of the whole feature space, while the learning margin in RPL can show the size of the whole feature space. Fig.\ref{fig:lamdbma} shows the variation trend of margin with $\lambda$. With the increase of  $\lambda$, the learned margin becomes smaller, and the limitation of $\mathcal{L}_o$ on feature space also tends larger. Different training data need different reasonable margins and space sizes to ensure that known and unknown can be classified correctly. Fig.\ref{fig:known} proves that the margin increases with the number of known classes with fixed lambda. Reasonable $\lambda$ can effectively control the interaction among known classes, by learning the more appropriate embedding space size. 
% More relevant visualizations of feature are provided in the Appendix \textcolor{red}{4}.

\noindent
\textbf{Experiments with Multiple Reciprocal Points.} 
Fig.\ref{fig:multi} shows the impact of increasing the number of reciprocal points on open set identification to detect rare unknown classes. We can observe that the performance of open set identification is relatively stable. The performance is slightly improved when employing 8 reciprocal points. Based on a good feature space distribution, increasing the number of reciprocal points for each known category can slightly improve the ability of the neural networks to distinguish unknown. However, too many reciprocal points will bring more training overhead and affect performance.

\noindent
\textbf{Combination of Reciprocal Point and Prototype.} 
The reciprocal point can push known classes to the periphery in the limited feature space, thereby improving the performance of the open set identification. We deem that the reciprocal point and the prototype are opposite and complementary. Here we combine the reciprocal point and the prototype, as \emph{RPL++}. In the training phase of RPL, we add prototypes and further train reciprocal points and prototypes by RPL and prototype loss \cite{yang2018robust}. As shown in Table. \ref{tab:osdi}, RPL++ achieves better performance than RPL. Moreover, the prototype further improves the performance of RPL on open long-tailed recognition in Table. \ref{tab:olr}. The reciprocal points form an excellent embedding space structure, then the prototype further narrows intra-class variance, to further divide the known classes and the unknown classes.

\section{Conclusion}
This paper proposes a new concept, \emph{Reciprocal Point}, which is the potential representation of the extra-class space corresponding to each known category. We also propose a novel learning framework towards open set learning. The approach introduces unknown information by reciprocal points, to optimize a better feature space to separate known and unknown. Comprehensive experiments conducted on multiple datasets demonstrate that our method outperforms previous state-of-the-art open set classifiers in all cases. We also publish a open long-tailed dataset, the Air-300, which is a challenging dataset to simulate natural data distribution for open set recognition and other visual tasks. 

\subsection*{Acknowledgments}
This work is partially supported by grants from the National Key R\&D Program of China under grant 2017YFB1002400, the Key-Area Research and Development Program of Guangdong Province under Grant 2019B010153002, and the National Natural Science Foundation of China under contract No.61825101, No.61702515 and No.U1611461.

\bibliographystyle{splncs04}
\bibliography{main}

\appendix

\section{Preliminaries}

Our framework identifies open set learning problem by integrating multiple binary classification tasks (\textit{one} \textit{vs.} \textit{rest}) into a multiclass recognition problem, by summing the expected risk defined in the paper category by category as:
\begin{equation}
	\mathop{\sum}_{k=1}^N \ \ \mathcal{R}_\epsilon(\psi_k, \ \mathcal{S}_k \cup \mathcal{O}^{pos}_k) + \alpha \cdot
	\mathcal{R}_o(\psi_k, \ \mathcal{O}^{neg}_k)
\label{Eqn:1}
\end{equation}
which can be reformulated as:
\begin{equation}
\mathop{\sum}_{k=1}^N \ \ \mathcal{R}_\epsilon(\psi_k, \ \mathcal{S}_k \cup \mathcal{O}^{pos}_k) + 
\alpha \cdot \mathop{\sum}_{k=1}^N \ \  \mathcal{R}_o(\psi_k, \ \mathcal{O}^{neg}_k)
\label{Eqn:2}
\end{equation}
where $\psi_k: \mathbb{R}^d \mapsto \{0, 1\}$ is a binary measurable prediction function, $\mathcal{S}_k$, $\mathcal{O}_k^{pos}$ and $\mathcal{O}_k^{neg}$ are known space, open positive space and infinite open negative space respectively. Minimizing the left part of plus sign in the Eq.\ \eqref{Eqn:2} can be viewed as training multi binary classifiers for each category, with summing, leading to a multi-class prediction function $f=\odot(\psi_1, \psi_2, \dots, \psi_N)$ for $N$-category classification, where $\odot(\cdot)$ is an integrating operation. Hereafter, we further replace $\psi_k$ with $f$ in the Eq.\ \eqref{Eqn:2} and get the following formulation as:
\begin{equation}
\mathop{\arg\min}_{f \in \mathcal{H}} \ \lbrace 
\mathcal{R}_\epsilon(f, \mathcal{D}_L) + \alpha \cdot 
\mathop{\sum}\nolimits_{k=1}^N\mathcal{R}_o(f, \mathcal{D}_U)
\label{Eqn:3}
\end{equation}
where $f: \mathbb{R}^d \mapsto \mathbb{N}$ is a measurable multi-class recognition function, $\mathcal{D}_L$ is all labeled data during training phase and $\mathcal{D}_U$ is potential unknown data. Minimizing the overall loss of the empirical classification risk and the open space risk simultaneously leads to be more distinguishable between known and unknown space. 

\section{RPL++}

The reciprocal point and the prototype are opposite and complementary. Here we simply combine the reciprocal point and the prototype. In addition to the reciprocal points, we added prototypes for known class $k$, denoted as $ \mathcal{M}^k = \{ m^k_i | i=1, \dots, C \} $ where $C$ represents the index of the prototypes in each known class. Prototypes are initialized to the center of each known class. In the training phase of RPL, we add prototypes for training by \emph{Prototype Loss} in \cite{yang2018robust}:
\begin{equation}
	\mathcal{L}_{pl}(x; \theta, \mathcal{M}^k) = \frac{1}{C} \cdot \sum^{C}_{i=1}{|| f_{\theta}(x) - m^k_i ||^2_2}
	\label{Eqn:plloss}
\end{equation}
The overall loss function of RPL + GCPL combines RPL and \emph{Prototype Loss} as:
\begin{equation}
	\mathcal{L}(x; \theta, \mathcal{P}, R, \mathcal{M}) = \mathcal{L}_c(x;\theta, \mathcal{P}) + \lambda \mathcal{L}_o(x; \theta, \mathcal{P}, R) + \beta \mathcal{L}_{pl}(x; \theta, \mathcal{M})
	\label{Eqn: RPLPL}
\end{equation}
where $\beta$ is a hyper-parameter of controlling the weight of \emph{Prototype Loss} (PL). PL loss is used to reduce intra-class variance. The reciprocal point forms an excellent embedding space structure, then the prototype further narrow the intra-class distance so as to further divide the known classes and the unknown classes.

\section{Implementation Details}
$\gamma$ is set as 0.5 in all training phases. Reciprocal points are initialized by the random normal distribution and each margin is initialized with zeros. In experiments for out-of-distribution, $ \lambda $ is set to 0.1 in the training stage and we utilize the output after global average pooling (GAP) of WRN-40-4 as the feature. For open set identification, we add a global average pooling after the final convolution layer of the encoder. The dimension of the reciprocal point is consistent with the output of the GAP. Expect experiments with multiple reciprocal points, each known class is assigned one reciprocal point for training. $\beta$ is set as 0.1 in the training phase by \emph{RPL++}. We train all deep neural networks for 100 epochs with batch size 128 and momentum 0.9. The learning rate starts at 0.01 and is dropped by a factor of 0.1 in the training progress every 30 epochs. Apart from SGD optimizer used in ImageNet-LT, neural networks are trained with Adam optimizer \cite{kingma2014adam}. 

\begin{figure*}[!tb]
	\centering
	\subfigure[$\lambda = 0.001$]{
		\begin{minipage}[t]{0.24\linewidth}
			\centering
			\includegraphics[width=\linewidth]{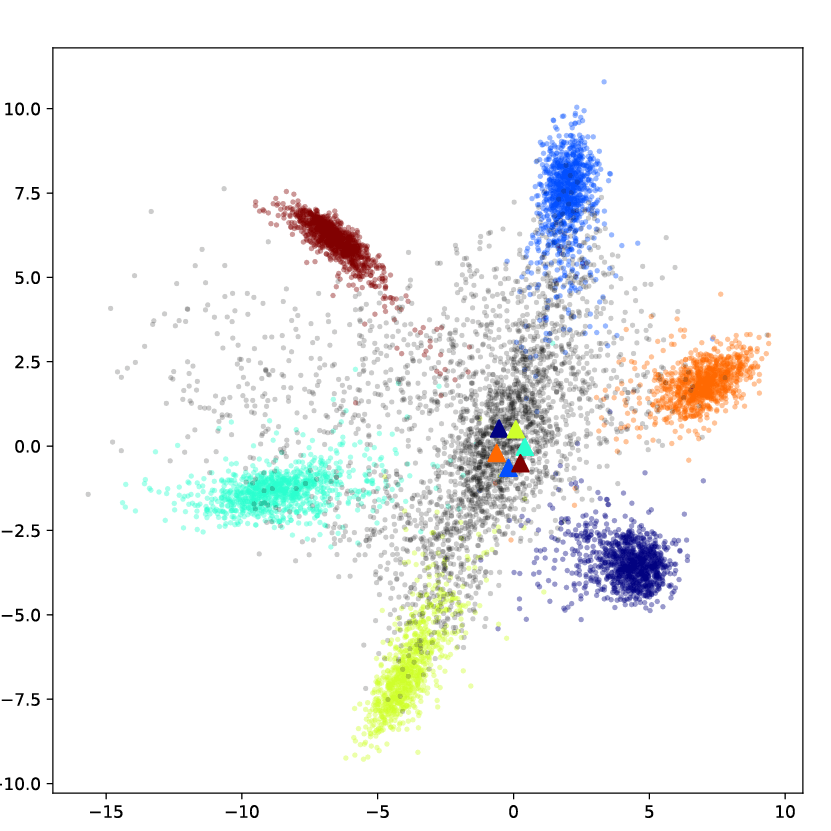}
			\label{fig:SPLR_0001}
		\end{minipage}%
	}%
	\subfigure[$\lambda = 0.01$]{
		\begin{minipage}[t]{0.24\linewidth}
			\centering
			\includegraphics[width=\linewidth]{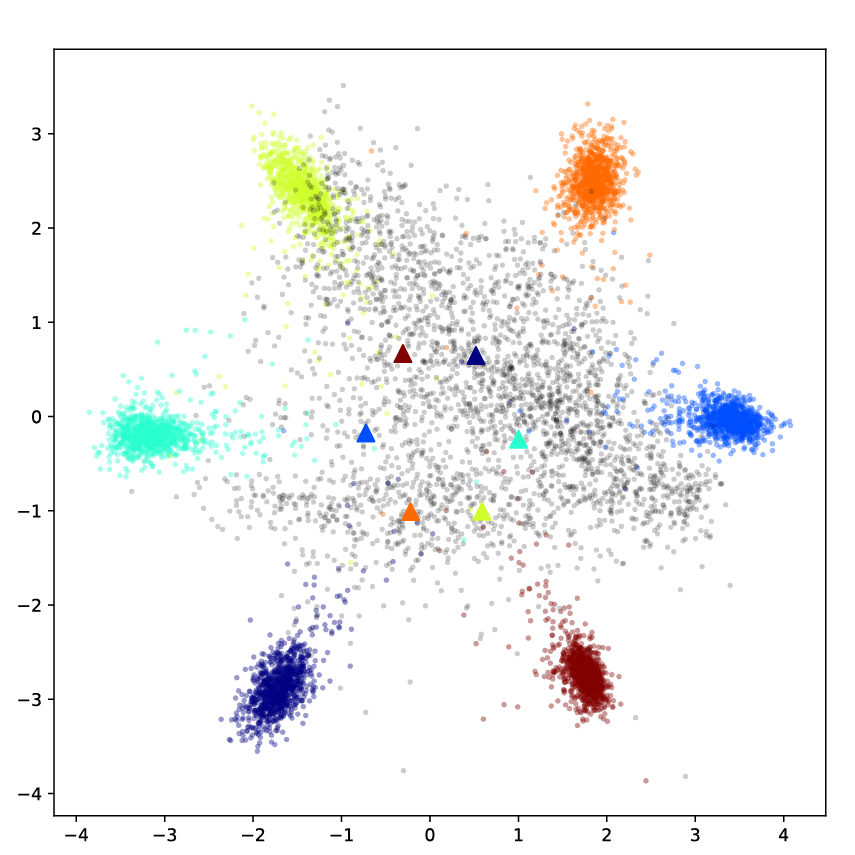}
			\label{fig:SPLR_001}
		\end{minipage}%
	}%
	\subfigure[$\lambda = 0.1$]{
		\begin{minipage}[t]{0.24\linewidth}
			\centering
			\includegraphics[width=\linewidth]{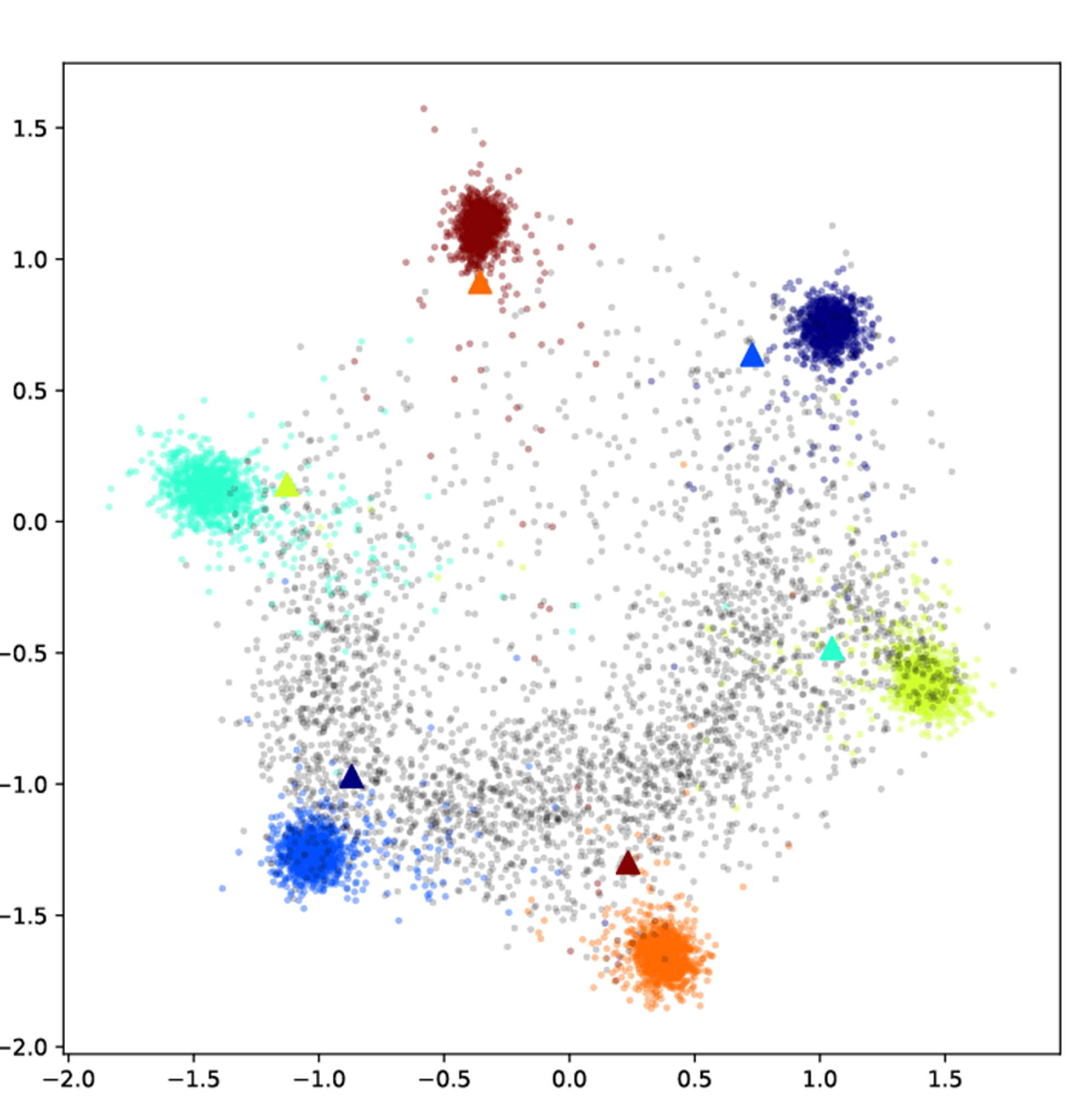}
			\label{fig:SPLR_01}
		\end{minipage}%
	}%
	\subfigure[$\lambda = 1$]{
		\begin{minipage}[t]{0.24\linewidth}
			\centering
			\includegraphics[width=\linewidth]{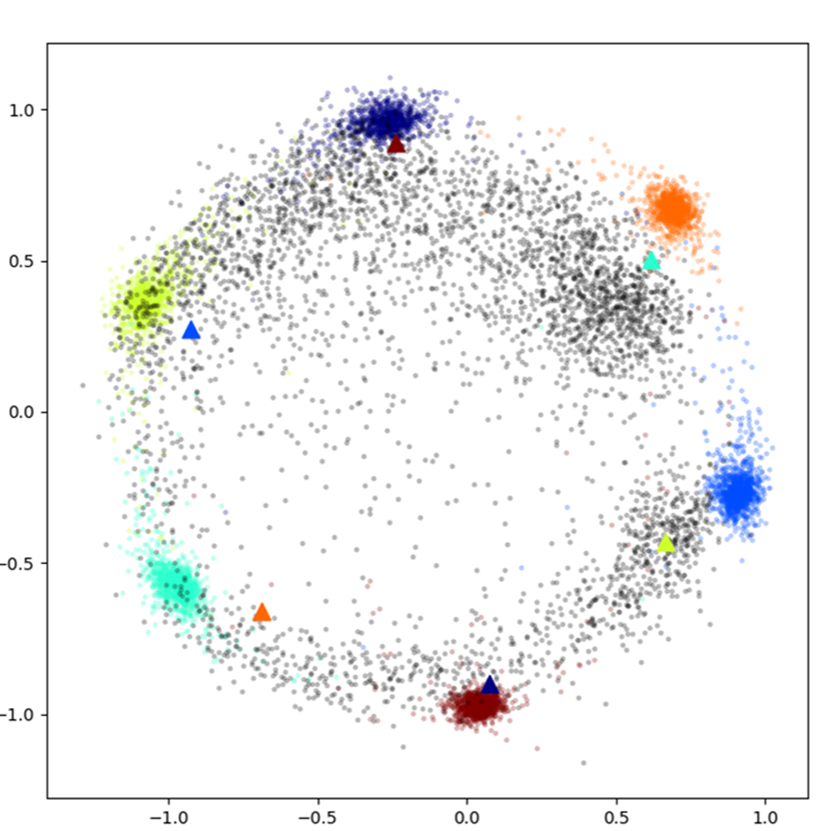}
			\label{fig:SPLR_1}
		\end{minipage}%
	}%
	\\
	\subfigure[$N = 2$]{
		\begin{minipage}[t]{0.24\linewidth}
			\centering
			\includegraphics[width=\linewidth]{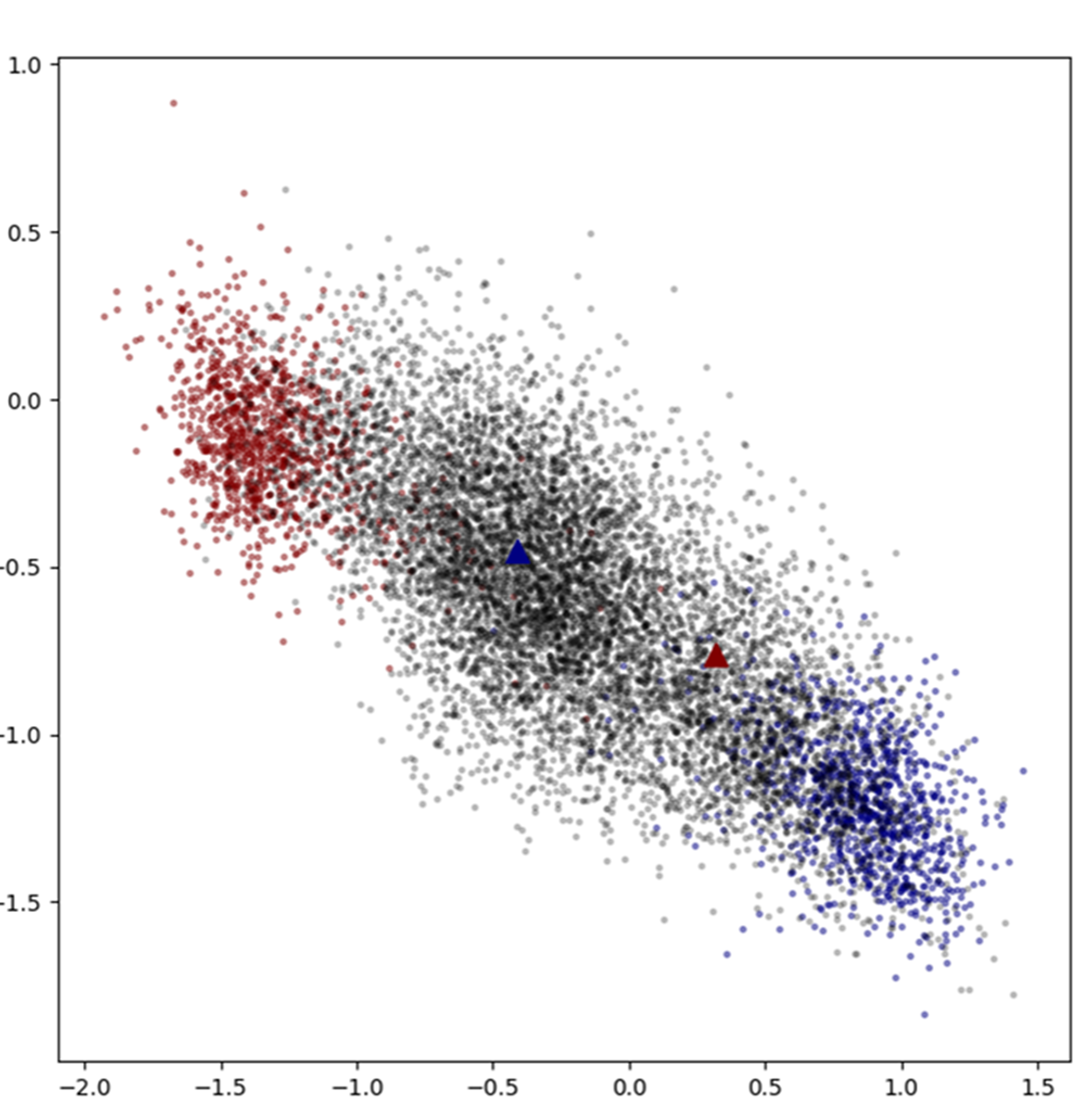}
			\label{fig:SPLR_2}
		\end{minipage}%
	}%
	\subfigure[$ N = 5 $]{
		\begin{minipage}[t]{0.24\linewidth}
			\centering
			\includegraphics[width=\linewidth]{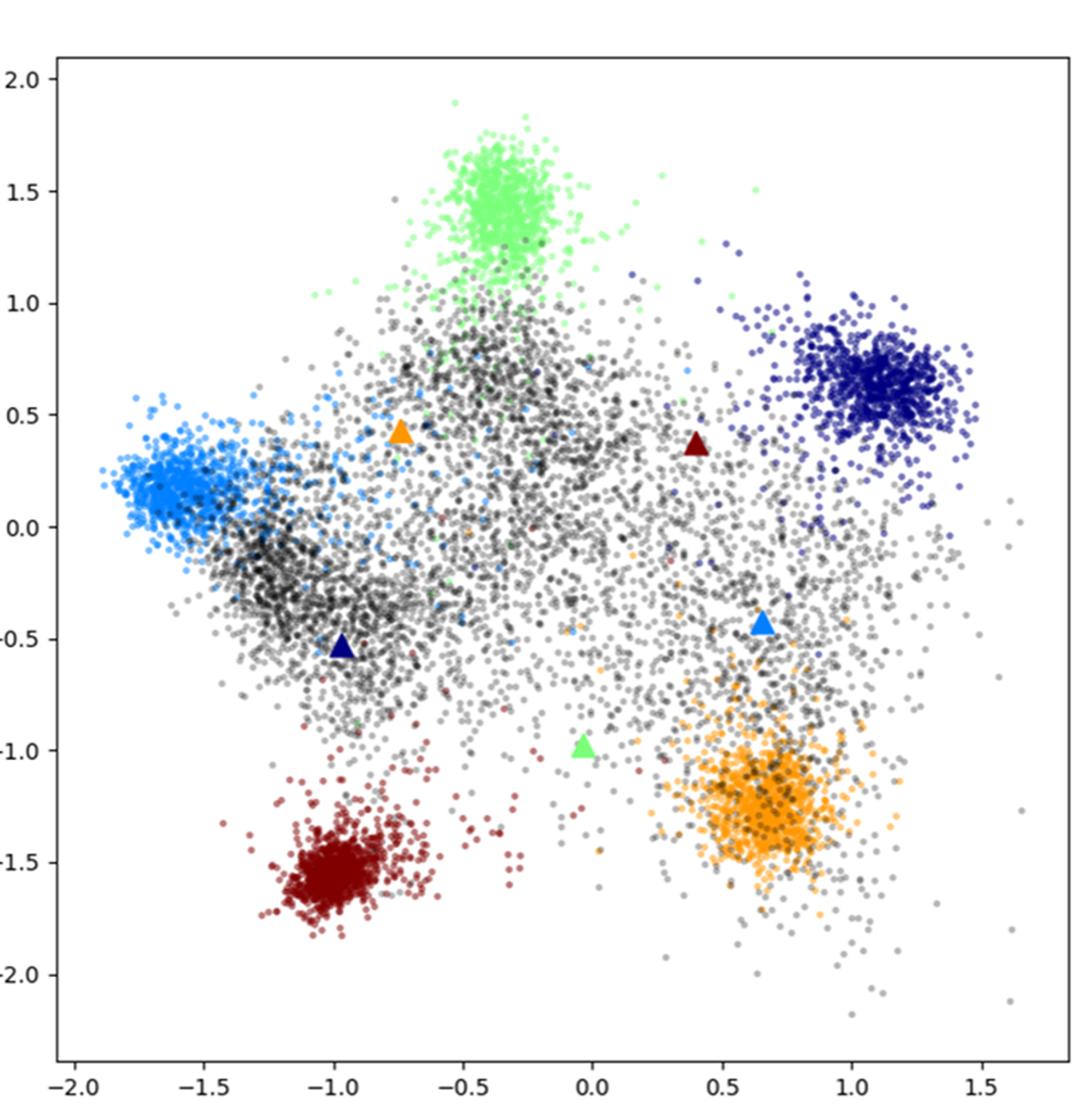}
			\label{fig:SPLR_5}
		\end{minipage}%
	}%
	\subfigure[$ N = 6 $]{
		\begin{minipage}[t]{0.24\linewidth}
			\centering
			\includegraphics[width=\linewidth]{figures/embedding/SPLR_6_01}
			\label{fig:SPLR_6}
		\end{minipage}%
	}%
	\subfigure[$ N = 9$]{
		\begin{minipage}[t]{0.24\linewidth}
			\centering
			\includegraphics[width=\linewidth]{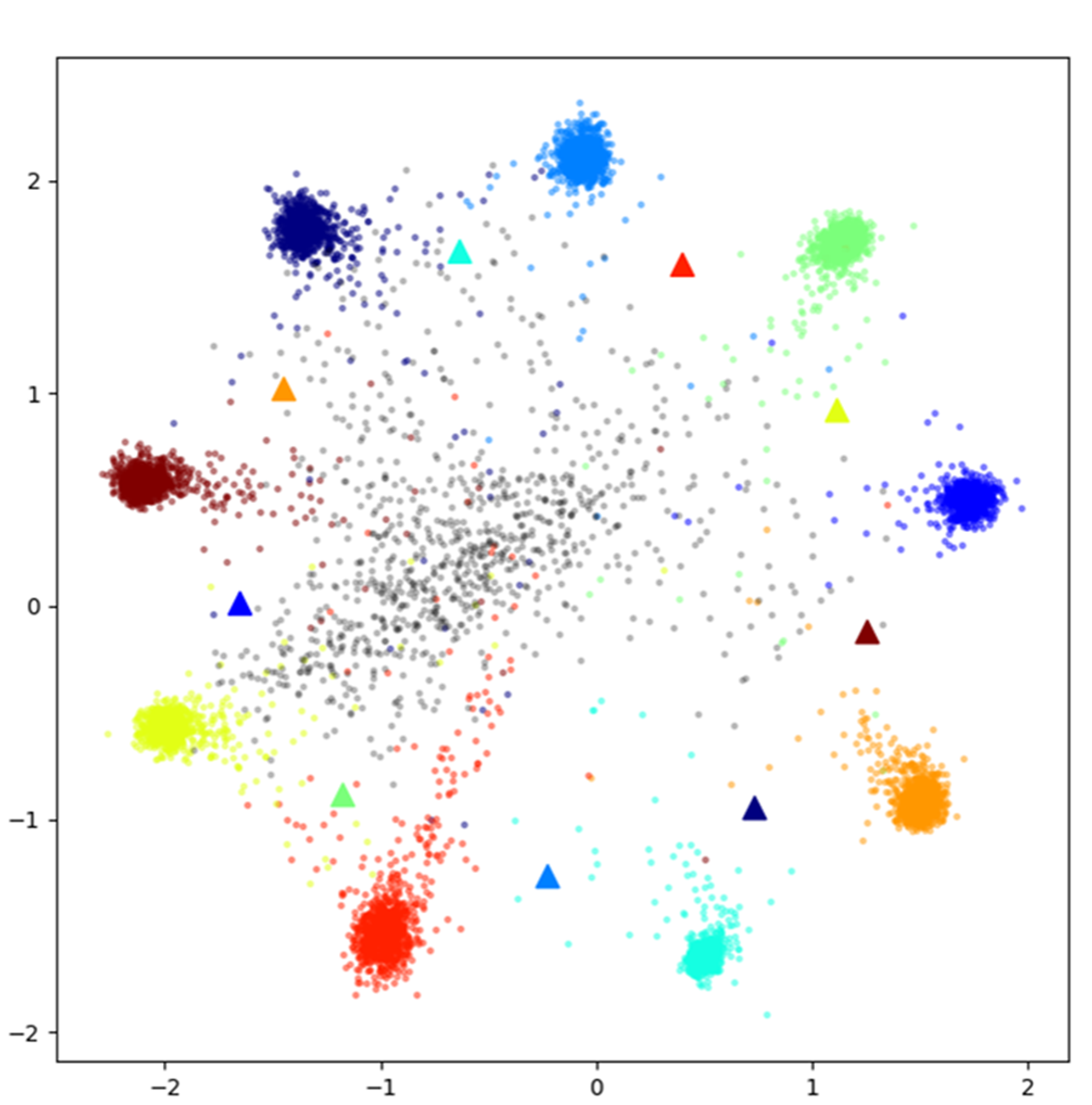}
			\label{fig:SPLR_9}
		\end{minipage}%
	}
	\caption{(a)-(d) The learned representations of RPL with different $\lambda$. The data is from MNIST by randomly sampling 6 known classes and 4 unknown classes. (e)-(h) The learned representations of RPL with different numbers of known classes. The data is from MNIST by randomly sampling $K$ known classes and $10-K$ unknown classes, where $\lambda = 0.1$.  Colored triangles represent the learned reciprocal points of different known classes.}
	\label{fig:embedding}
\end{figure*}

\begin{figure*}[!tb]
	\centering
	\subfigure[$ epoch = 0 $]{
		\begin{minipage}[t]{0.45\linewidth}
			\centering
			\includegraphics[width=\linewidth]{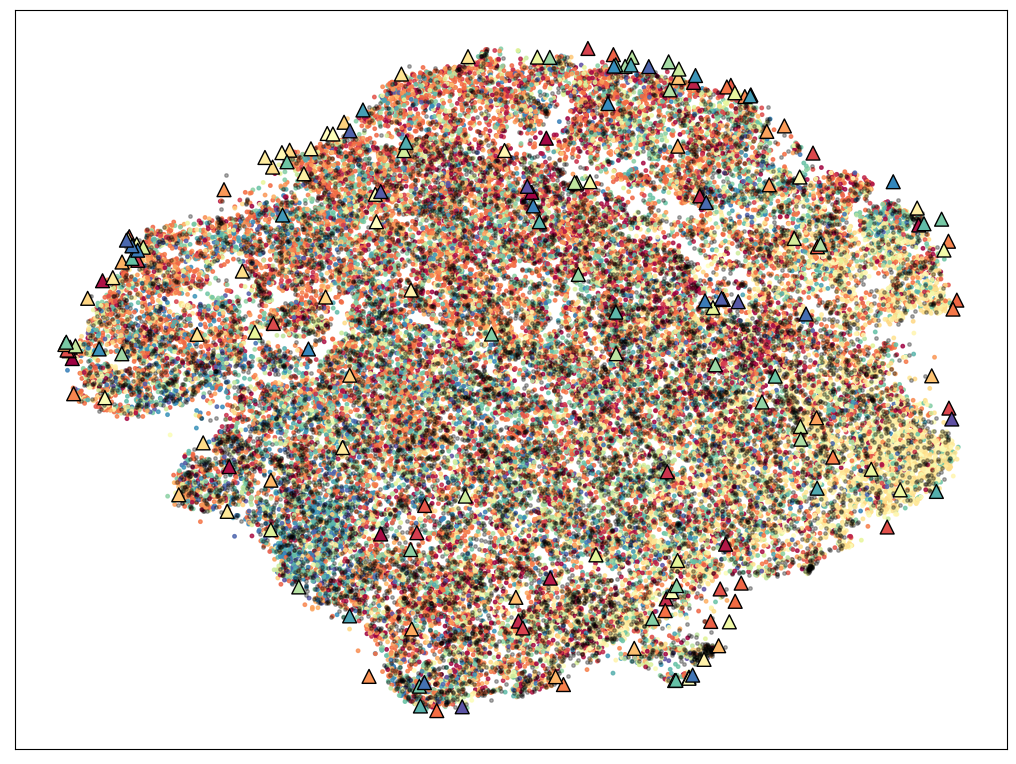}
			\label{fig:RPL_1}
		\end{minipage}
	}
	\subfigure[$ epoch = 100 $]{
		\begin{minipage}[t]{0.45\linewidth}
			\centering
			\includegraphics[width=\linewidth]{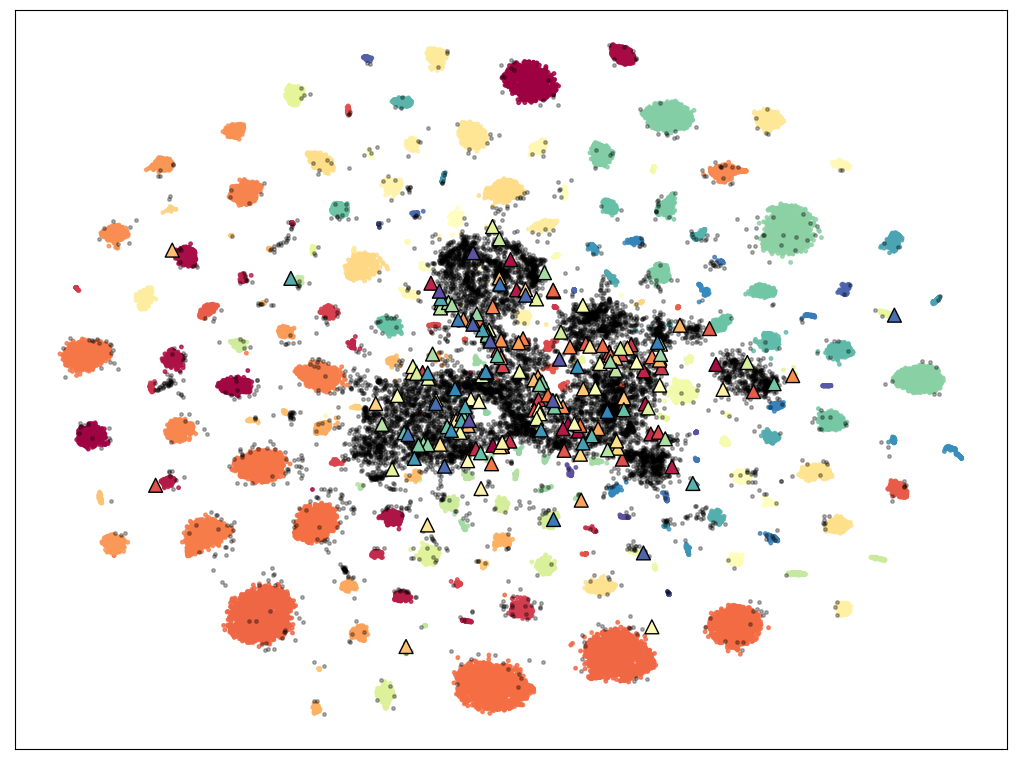}
			\label{fig:RPL_100}
		\end{minipage}
	}
	\caption{(a)-(b) The learned representations of RPL in different training stages on Air-300. All classes are divided into two parts with 180 known classes for training and 120 novel unknown classes for testing respectively. Colored triangles represent the learned reciprocal points of different known classes.}
	\label{fig:air}
\end{figure*}

\section{Additional Results}

\subsection{Ablation Study}

As shown in Fig.\ref{fig:embedding}, the reciprocal points are relative to the corresponding known classes in embedding space. The unknown samples are limited to the internal bounded space and all known classes are distributed around the periphery of the space. The size of the entire embedding space is limited by $\mathcal{L}_o$. However, with the increase in known classes, the entire embedding space should become larger for robust classification. Fig.\ref{fig:embedding} (e)-(h) shows the distribution of features learned under different numbers of known classes. It can be seen through the axes that the known class distribution becomes wider with more known classes. Meanwhile, the margin $R$ is also grown with more known classes. This phenomenon proves the rationality of the spatial distribution learned for multiclass.

As shown in Fig. \ref{fig:RPL_1} and \ref{fig:RPL_100}, we provide the change of embedding features generated by t-SNE \cite{maaten2008visualizing} on the Air-300. Compared with the initial features of neural network in Fig.\ref{fig:RPL_1}, the features of known and unknown classes are more obviously separated through RPL. We can notice that the reciprocal points are almost in the embedding space of unknown classes in Fig.\ref{fig:RPL_100}, which also proves that RPL can introduce the unknown information in the training process. It also produces the phenomenon that unknown classes are in the middle and the known category is distributed in outer space, which also proves the effectiveness of RPL for multiclass.

\begin{table*}[!tb]
	\caption{The results for long-tailed recognition.}
	\centering
	\label{tab:ltr}
	\setlength{\tabcolsep}{2mm}{
	\begin{tabular}{cccc}
		\toprule
		\textbf{Methods} & \textbf{All*} & \textbf{Head (80\%) categories} & \textbf{Tail (20\%) categories} \\
		\midrule
		Softmax & 85.7 & 86.5 & 69.3 \\
		GCPL & 84.5 & 85.4 & 66.2 \\
		RPL & \textbf{88.8} & \textbf{89.5} & \textbf{74.0} \\
		\bottomrule
	\end{tabular}}
\end{table*}

% \noindent
\subsection{The Long-tailed Recognition}
The long-tail experiments are to simulate real-world distribution so as to prove the effectiveness of RPL in practical application. The results prove that RPL is able to distinguish unknown categories in long-tail scenario. We further conduct classification experiments on the head (80\%) categories and tail (20\%) categories of Air-300 by using the same setting in the Sec 4.2. According to the results in Table \ref{tab:ltr}, RPL is able to recognize unknowns while maintaining similar or even better classification performance. 

\begin{table*}[!tb]
	\caption{F1-scores against varying Openness with different baselines.}
	\centering
	\label{tab:openess}
	\setlength{\tabcolsep}{3mm}{
	\begin{tabular}{ccccccc}
		\toprule
		\textbf{Openness} & \textbf{0.18} & \textbf{0.24} & \textbf{0.3} & \textbf{0.36} & \textbf{0.42} & \textbf{0.49} \\
		\midrule
		Softmax & 59.0 & 49.5 & 41.1 & 33.4 & 26.8 & 20.2 \\
		GCPL & 58.6 & 49.2 & 40.7 & 33.1 & 26.5 & 20.0 \\
		RPL & 60.4 & 51.0 & 42.5 & 34.7 & 27.9 & 21.2 \\
		RPL++ & \textbf{60.6} & \textbf{51.2} & \textbf{42.7} & \textbf{34.9} & \textbf{28.1} & \textbf{21.4} \\
		\bottomrule
	\end{tabular}}
\end{table*}

\subsection{Open Set Recognition}
Table \ref{tab:openess} shows the F-measure (or F1-scores) \cite{powers2011evaluation} trend under different Openness \cite{scheirer2013toward} in cifar100 averaged over ten randomized trials. In each trial, 15 categories are randomly selected as knowns and unknowns are randomly selected from the rest categories according to the Openness. We use ResNet18 as backbone networks and use the threshold of 0.1 for all methods. To evaluate all methods at a uniform threshold, all outputs are normalized by softmax function. The results show that RPL and RPL++ are superior to Softmax and GCPL as the Openness increasing.

\begin{figure*}[!b]
	\centering
	\includegraphics[width=0.98\linewidth]{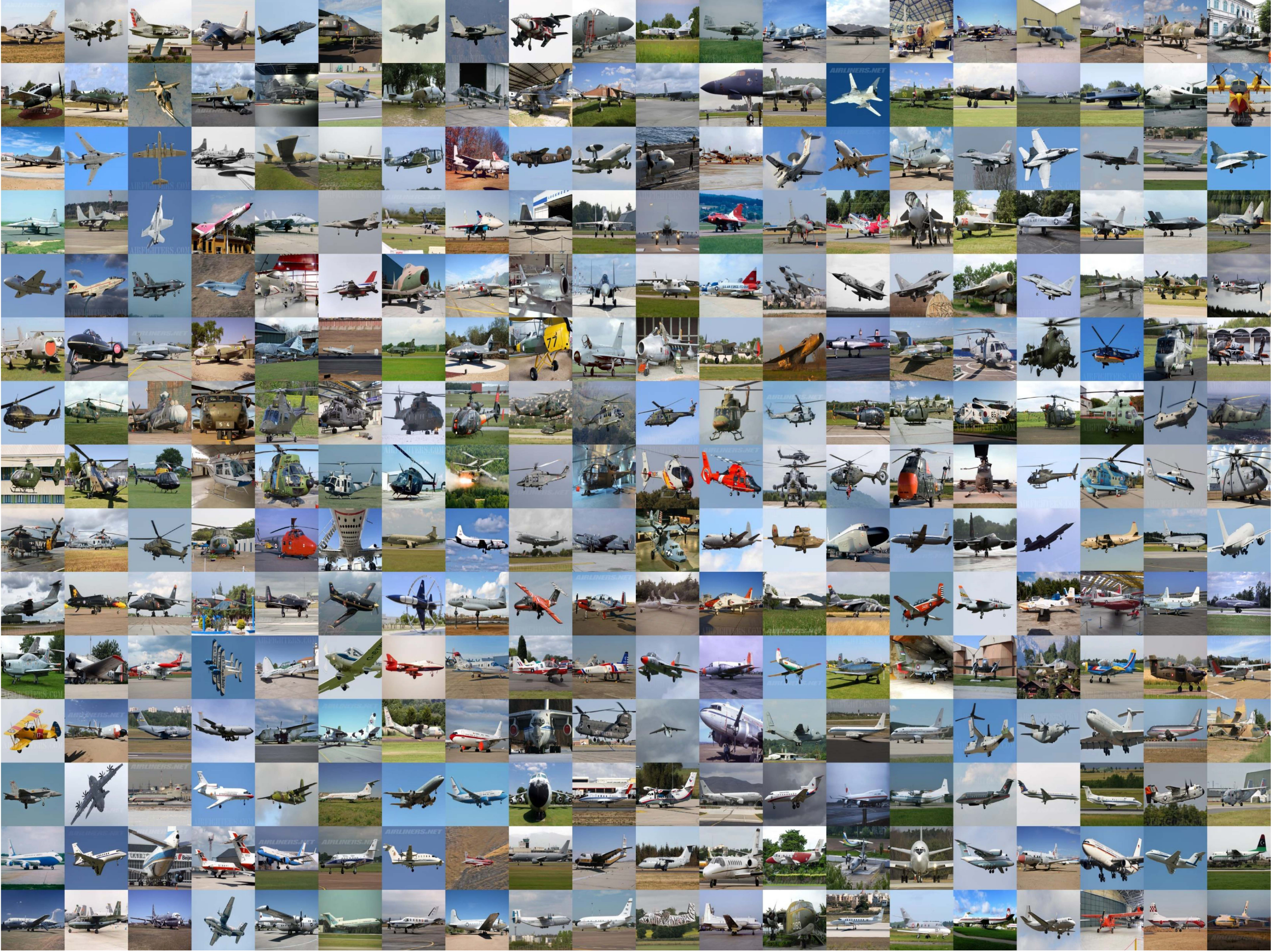}
	\caption{The overview of Air-300.}
	\label{fig:air300}
\end{figure*}

\end{document}